\documentclass[11pt]{article}

\usepackage[final]{acl}

\usepackage{times}
\usepackage{latexsym}
\usepackage[T1]{fontenc}
\usepackage[utf8]{inputenc}
\usepackage{microtype}
\usepackage{inconsolata}
\usepackage{graphicx}
\usepackage{setspace}
\usepackage{booktabs}
\usepackage{float}
\usepackage{caption}
\usepackage{capt-of}
\usepackage{amsmath}
\usepackage{cleveref}

\usepackage{placeins}
\usepackage{tikz-dependency}
\usepackage{fontawesome}
\usepackage{supertabular}
\usepackage{multicol}

\newcommand{\xtk}[2]{x_{#1}^{(#2)}}

\newcommand{\wtk}[2]{w_{#1}^{(#2)}}

\newcommand{\utt}{\mathbf{u}}

\newcommand{\uttab}[2]{\utt_{#1:#2}}

\newcommand{\probP}{P}

\usepackage[dvipsnames]{xcolor}

\newcommand{\CriticalWord}{\textcolor{black}{{\texttt{CriticalWord}}}}
\newcommand{\Preamble}{\textcolor{black}{{\texttt{Preamble}}}}
\newcommand{\Predicate}{\textcolor{black}{{\texttt{Predicate}}}}
\newcommand{\Intervening}{\textcolor{black}{{\texttt{Intervening}}}}

\newcommand{\Plausible}{\textcolor{ForestGreen}{\texttt{Plausible}}}
\newcommand{\NeighborGP}{\textcolor{orange}{\texttt{Neighbor-GP}}}
\newcommand{\UnrelatedGP}{\textcolor{RoyalBlue}{\texttt{Unrelated-GP}}}
\newcommand{\LateError}{\textcolor{RoyalPurple}{\texttt{Late-Error}}}
\newcommand{\Typo}{\textcolor{Gray}{\texttt{Typo}}}

\title{Readers make targeted regressions to plausible errors in \\ reanalysis of ``noisy-channel garden-path'' sentences}

\author{Thomas Hikaru Clark \: \: \: Roger Levy \: \: \: Edward Gibson \\
        MIT Brain and Cognitive Sciences \\ 43 Vassar Street, Cambridge MA, USA \\ 
        \small{
   \textbf{Correspondence:} \href{mailto:thclark@mit.edu}{thclark@mit.edu}
 }
 }

\begin{document}

\maketitle

\begin{abstract}
A key question in psycholinguistics is how inferences about the meaning of linguistic input unfold incrementally a comprehender's mind. 
In this work, we study reading dynamics for ``noisy-channel garden-path'' sentences, which temporarily appear well-formed but feature late-appearing violations of expectation that can be resolved not by inferring an alternative syntactic structure, but by inferring the presence of an error. 
We find evidence for targeted regressions -- eye movements towards regions that are promising loci of possible errors in light of later-arriving information, showing patterns consistent with the posterior inferences of a model of noisy-channel processing with reanalysis. 
We discuss the implications of these findings for theories of noisy-channel language comprehension and information-theoretic explanations of reading dynamics. 
\end{abstract}

\section{Introduction}
\label{sec:intro}

A key question in psycholinguistics is how inferences about the meaning of linguistic input unfold over time in the mind of a comprehender. 
Natural language contains \textbf{ambiguity}: a given input can be compatible with multiple latent grammatical structures (e.g., \textit{I watched the hiker with the binoculars}), sound sequences can be ambiguous between different lexical items, and pronouns can be ambiguous between different referents.
Language comprehension is also constrained by \textbf{incrementality}: linguistic input is typically processed unit by unit in a linear sequence. Under strict incrementality, information about later words in a sentence cannot influence the initial processing of earlier words. 
To handle ambiguity, an incremental comprehender needs to either maintain uncertainty regarding an ambiguous word or structure until disambiguating information is reached, or commit to one interpretation, even if it ultimately proves incorrect. 

Sentences in which incremental processing is led astray tend to elicit processing difficulties in humans; these examples provide insights into the nature of sentence comprehension. 
In \textbf{garden-path sentences}, an initially likely incremental syntactic parse of a sentence is rendered incompatible by later disambiguating material, potentially forcing comprehenders to \textbf{reanalyze} the sentence as having an alternative structure. 
An open question is whether this garden-path effect extends not just to syntactic reanalysis, but to a broader notion of error-aware reanalysis, where readers take into account possible errors and noise in the linguistic signal, in keeping with \textbf{noisy-channel} theories of comprehension \citep{levyNoisyChannelModelHuman2008, gibsonRationalIntegrationNoisy2013}.

To probe the nature of reanalysis during comprehension, \textbf{regressive eye movements} are particularly informative. During reading, readers do not simply advance their gaze sequentially from word to word, but often revisit earlier words, suggesting a shift in attention to, and possible reanalysis of, previously processed material. 
In this study, we investigate whether readers make targeted regressions to the locations of explainable errors.
Distinct from syntactic analysis, this evaluates whether reading behavior is sensitive to possible production errors in the linguistic input, a prediction of noisy-channel theories.
This paper attempts to answer two research questions. 
\textbf{Q1}: Does reanalysis during language processing extend to possible errors and not just syntactic ambiguity?
\textbf{Q2}: Can an algorithmic-level model of noisy-channel language processing predict patterns of reading regressions for anomalous sentences?
To foreshadow our findings, we find evidence for targeted regressions to the location of likely word substitution errors, and demonstrate that this is consistent with the posterior predictions of a noisy-channel inference model. 

\subsection{Incremental processing and garden-paths}

Much of the research on reading behavior in psycholinguistics has been focused on incremental processing. According to surprisal theory \citep{haleProbabilisticEarleyParser2001, smithEffectWordPredictability2013, levyExpectationbasedSyntacticComprehension2008}, the processing difficulty of a given linguistic unit, e.g. a word, is proportional to its surprisal (or negative log probability) in context. 
Many experimental reading paradigms, such as self-paced reading and the Maze task, only provide incremental reading measures \citep{aaronsonPerformanceTheoriesSentence1976,mitchellEffectsContextContent1978,forsterMazeTaskMeasuring2009}.

Classic garden-path sentences provide a clear example of incremental processing difficulty induced by the ambiguity present in language \citep{beverCognitiveBasisLinguistic1970, frazierComprehendingSentencesSyntactic1979, paapeEstimatingTrueCost2022}. 
For example, in the sentence \textit{The bird perched on the branch sang sweetly}, the word \textit{perched} initially invites being parsed as the main verb; however, the word \textit{sang}, when reached, rules out this parse. 
Crucially, the existence of a garden-path effect suggests that comprehenders do not necessarily track all possible grammatical structures consistent with a given input prefix, but may instead greedily commit to an initially likely hypothesis, resulting in detectable slowdowns when disambiguating material is encountered. 



\subsection{Regressive eye movements in reading}

Human reading behavior is not always strictly incremental. 
Regressive eye movements, or \textbf{regressions}, are common --- it is estimated that 5-20\% of saccades are regressive \citep{raynerEyeMovementsReading1998}. 
While some regressions may be driven by low-level spatial constraints, such as the physical layout of text \citep{mitchellAccountingRegressiveEyemovements2008}, preventing readers from making regressions was shown to harm comprehension \citep{schotterDontBelieveWhat2014}; meanwhile, garden-path sentences elicit more regressions than control materials, which supports the idea that structural ambiguity leads to re-reading.

The Selective Reanalysis hypothesis \citep{frazierMakingCorrectingErrors1982} proposes that reading regressions represent readers' attempts to reanalyze the syntactic structure of earlier material.
Under this account, readers can make errors in incremental parsing by committing to the wrong syntactic parse, which causes later violations of expectations; readers then tend to regress to the location of the parsing error to re-parse the input. This provides one explanation of why reading regressions are especially likely to initiate at the disambiguating point of garden-path sentences. 
Recent work, however, has questioned the validity of the Selective Reanalysis Hypothesis, finding that garden-path sentences were often misunderstood and that re-reading did not improve comprehension \citep{christiansonRetracingGardenpathNonselective2024}.
\citet{paapeDoesLocalCoherence2021} found only weak and inconclusive support for targeted regressions to critical context words in sentences with late-breaking anomalies, and \citet{paapeConsciousRereadingConfirmatory2022} found support for confirmatory but not reanalytical regressions in syntactic garden-path sentences, using a bidirectional self-paced reading paradigm. Meanwhile, a recent large-scale eye-tracking study of re-reading behavior by \citet{timkeyEyeMovementsReveal2025} found evidence for targeted regressions to critical regions in syntactically challenging sentences. 

\citet{wilcoxInformationtheoreticAnalysisTargeted2024} perform an information-theoretic analysis of regressions, and find that high \textbf{pointwise mutual information} (PMI) between pairs of words in a sentence are associated with regressions during reading. 
High PMI signifies that two words are predictive of each other (e.g. \textit{I gave the \textbf{dog} a \textbf{bone}}), while low PMI signifies that two words tend to appear together \textit{less} than one would expect by chance (e.g. \textit{I gave the \textbf{dog} a \textbf{knife}}). 
This result is taken as evidence for a \textbf{reactivation account} of regressions, as opposed to a \textbf{reanalysis account}: readers tend to revisit words which are information-theoretically congruent, not incongruent, with the current word. 
The reactivation account is consistent with work finding that reading regressions are predicted by syntactic dependencies within a sentence \citep{lopopoloDependencyParsingYour2019}, which tend to exist between pairs of words with high mutual information \citep{futrellInformationtheoreticLocalityProperties2019a}. 

A question left open by past work is how regressions manifest in sentences that contain production errors (e.g. word substitutions), as opposed to syntactic ambiguity. Given that errors in printed texts are relatively rare, one might expect to find little evidence for the reanalysis account in reading of naturalistic language (the focus of most past studies), or that this evidence would be dominated by the much more common case of regressions for reactivation --- a possibility directly raised directly by \citet{wilcoxInformationtheoreticAnalysisTargeted2024}. 
Word substitutions, for example, preserve dependency structure while creating pairs of words in a dependency relation which nevertheless have low PMI; it is currently unclear whether human readers would be more or less likely to make regressive eye movements between such a pair of words.

\subsection{Algorithmic accounts of noisy-channel language processing}

The noisy-channel theory of language processing proposes that comprehenders arrive at non-veridical interpretations of sentences when more plausible alternatives exist, rationally integrating both the prior probability of different intended sentences and the error likelihood \citep{levyNoisyChannelModelHuman2008, gibsonRationalIntegrationNoisy2013}. 
Recent work has modeled how sentence processing for anomalous linguistic input may unfold over time for sentences with strong violations of incremental expectations, like \textit{The storyteller could turn any story into an amusing \textbf{antidote}} \citep{ryskinERPIndexRealtime2021, liHeuristicInterpretationRational2023, liInformationtheoreticModelShallow2024}.

\citet{clarkModelApproximateIncremental2025} propose a model with specific algorithmic commitments regarding both the incremental processing and reanalysis of possibly noisy utterances. In this model, Sequential Monte Carlo (SMC) inference is employed to approximate the posterior distribution over intended messages, conditional on an observed noisy string. SMC is an approximate Bayesian inference algorithm for sequential observations, which naturally instantiates a tradeoff between the number of particles and the exactness of inference: as the number of particles approaches infinity, the SMC approximation approaches the true posterior \citep{lewSequentialMonteCarlo2023, doucetIntroductionSequentialMonte2001, naessethElementsSequentialMonte2024}. 
In past work, the number of SMC particles is used as a proxy for cognitive resources \citep{levyModelingEffectsMemory2008,clarkResourceRationalNoisyChannelLanguage2025}. 
This computational approach therefore provides an explanation for how human comprehenders can be approximately rational, despite the intractability of computing exact posterior inferences regarding a large space of alternative interpretations of a noisy sentence. 

This model is by default incremental, processing words in sequential order and updating the posterior distribution over latent variables at each time step. 
At the same time, the model supports reanalysis of earlier commitments via \textbf{rejuvenation} moves that propose changes to earlier commitments (e.g., inferring that an earlier word was actually an error).  
Unlike \citet{levyModelingEffectsMemory2008}, which treats the input string as veridical and performs inference over the uncertainty in latent syntactic structures, this noisy-channel model treats the input string as possibly non-veridical and performs inference over possible alternative \textit{intended} sentences. 
In this work, we use this computational model to generate posterior inferences for possibly noisy experimental stimuli, yielding predictions of how the experimental conditions may systematically differ from each other (\Cref{sec:methods-model}).
In particular, the model identifies words which are likely to be errors in light of later-arriving information, even if they initially appear perfectly well-formed using only previous context. 

\subsection{``Noisy-channel garden-path'' sentences as a key testbed for theories of processing}
\FloatBarrier

\begin{table*}[htb]

	\caption{Example stimuli --- plausible and garden-path sentences.}
	\centering
    \small
	\begin{tabular}{l|llll}
		\toprule
		\textbf{Condition}     & \textbf{Preamble}  & \textbf{CriticalWord} &  \textbf{Intervening} & \textbf{Predicate} \\
		\midrule
		\Plausible{}     & The boy   & kicked  & the big round & ball into the net.  \\
            \Plausible{}     & The boy   & licked  & the big round & lollipop with delight.    \\
		\NeighborGP{}   & The boy   & licked  & the big round & ball into the net.    \\
        \NeighborGP{}   & The boy   & kicked  & the big round & lollipop with delight.  \\
		\bottomrule
	\end{tabular}
	\label{tab:example-materials-main}
\end{table*}

A useful case for testing theories of reanalysis in reading is a sentence which elicits a garden-path effect that can be resolved not by syntactic reanalysis but by hypothesizing a production error at an earlier part of the sentence. An example of such a \textbf{``noisy-channel garden-path''} sentence is \textit{The boy licked the big round ball into the net} (full example in \Cref{tab:example-materials-main}).
When such a sentence is being incrementally processed, the \Predicate{} is semantically, rather than syntactically, incongruous with earlier context. This incongruity, paired with the form-based similarity of the word \textit{licked} to the more globally congruous word \textit{kicked}, potentially invites an error-correction inference on the part of a comprehender.
Each implausible sentence can be directly compared to a plausible counterpart which is identical up until the \Predicate{}.

Critically, no purely incremental processing algorithm (e.g., autoregressive language model surprisal) can make different predictions regarding regressive reading specifically to \CriticalWord{} in the matched \Plausible{} and \NeighborGP{} sentences (see \Cref{tab:example-materials-main}). 
Incremental language model surprisal \textit{does} predict differences in behavior at the \Predicate{}, and one plausible hypothesis is that when a threshold of surprisal is reached, some form of repair or reanalysis is triggered. 
However, to know specifically \textit{where} a comprehender should focus their re-reading effort, we need an algorithm that instantiates reanalysis.
Unlike in traditional garden-path sentences, these noisy-channel garden-paths have a specific location that can be hypothesized as a production error.
Three plausible accounts are described below. 


\textbf{A: Purely incremental account.}
Incremental features capture all explainable variation in reading times. When encountering a semantic violation at the \Predicate{}, the reader either covertly performs reanalysis without an eye-movement correlate, or accepts the semantically incongruous but syntactically valid \Predicate{} without reanalysis. 
In either case, we expect a slowdown at the \Predicate{}, but this account does not predict regressions to earlier words. 

\textbf{B: Surprisal triggers non-targeted re-reading.}
When surprising material is reached, readers are more likely to initiate a regression to earlier material, without specifically targeting the location of likely errors. 
A simple version of this account is that readers simply re-read the sentence from the beginning. 
Under this account, the \CriticalWord{} in the \NeighborGP{} condition in \Cref{tab:example-materials-main} would not attract more re-reading than other regions in the sentence.

\textbf{C: Surprisal triggers targeted re-reading.}
This account predicts that when surprising material is reached, readers are more likely to initiate a regression targeted to the location of likely errors in the sentence. 
One theory that predicts targeted re-reading is an algorithmic-level model of noisy-channel comprehension, which proposes multiple possible cues that may identify an earlier word in a sentence as being a fruitful location for reanalysis. 
Broadly, these cues include the linguistic \textbf{prior} (how probable a word is in context) and the error \textbf{likelihood} (how likely a particular error is, according to a comprehender's mental model of errors). The challenge for a reader is to identify earlier words which (a) have an alternative interpretation that is a near neighbor in the space of errors and (b) this alternative leads to a more plausible (higher-prior) sentence.
Additionally, this requires a linking hypothesis, which is that reanalysis of a word is mediated by additional eye movements towards that word after it has been read, and the ensuing additional reading duration. 

\section{Methodology}
\label{sec:methods}

We conduct a reading experiment aimed at answering our key research questions. 
The experimental setup and analysis plan were preregistered via OSF: \url{https://osf.io/qtnxa}. 

\subsection{Materials}
\label{sec:methods-materials}

We systematically manipulate sentences to test reading behavior for noisy-channel garden path sentences and a variety of control conditions (\Cref{tab:example-materials-main,tab:example-materials-controls}). 
All target materials consist of the following regions: \Preamble{}, \CriticalWord{}, \Intervening{}, and \Predicate{}. 
We generated 36 items, each of which appears in 5 conditions. 
Additionally, 36 filler items containing no errors were shown. 
More examples of experimental materials are shown in \Cref{sec:extra-examples}.

\begin{table}
\centering
\footnotesize
\begin{tabular}{l|ll}
    \toprule
    \textbf{Condition}   & \textbf{CriticalWord} & \textbf{Predicate} \\
    \midrule
        \Typo{}   & kjcked  & ball into the net.  \\
        \Typo{}    & ljcked  & lollipop with delight.  \\
        \UnrelatedGP{}   & read   & ball into the net.  \\
        \UnrelatedGP{}    & read   & lollipop with delight.  \\
        \LateError{}  & kicked   & breath after the run.  \\
        \LateError{}   & licked   & breath after the run.  \\
    \bottomrule
\end{tabular}
\caption{Additional control stimuli for one item. All variants of an item share the same \Preamble{} and \Intervening{} material: ``The boy'' and ``the big round''.}
\label{tab:example-materials-controls}
\end{table}

Within each condition, items are counterbalanced by having two versions of \CriticalWord{}; this leads to 2 variants of each item per condition. The \NeighborGP{} condition is formed from the \Plausible{} condition by simply swapping the \Predicate{} regions of the two variants. This leads to a tightly controlled comparison where lexical items are completely balanced, with plausibility depending only on the identity of the \Predicate{}. 
For example, both prefixes \textit{The boy licked the big round...} and \textit{The boy kicked the big round...} appear in both the Plausible and Garden-Path conditions, depending only on the value of \Predicate{} (\textit{ball into the net} or \textit{lollipop with delight}).
Crucially, \CriticalWord{} in the two variants are orthographic neighbors of each other, e.g. \textit{lick} and \textit{kick}, providing a way of assessing sensitivity to error likelihood.

The \UnrelatedGP{} condition provides a comparison where the \CriticalWord{} region is an unrelated (non-neighbor) lexical item. These sentences can also be interpreted by positing a word substitution error at the \CriticalWord{} region, but unlike in the \NeighborGP{} condition, the orthographic form of \CriticalWord{} provides no cue towards a more plausible word.
In the case of \Typo{} errors, \CriticalWord{} is a non-word. 
This means that the sentence does not contain any temporary ambiguity, and thus no garden-path effect.
In the \LateError{} condition, \Predicate{} is incoherent with the earlier parts of the sentence. Thus, while the material up to the \Predicate{} is matched by sentences in the \Plausible{} condition, the sentence resists reanalysis via a simple word substitution error.

\subsection{Reading time data collection}
\label{sec:methods-reading}


To gather reading regression time data experimentally, we employ the Mouse Tracking for Reading (MoTR) paradigm \citep{wilcoxMouseTrackingReading2024}.
In other reading paradigms, such as unidirectional self-paced reading (SPR) and Maze, regressions are not possible \citep{aaronsonPerformanceTheoriesSentence1976,mitchellEffectsContextContent1978,futrellNaturalStoriesCorpus2021,forsterMazeTaskMeasuring2009}. Meanwhile, bi-directional SPR found evidence for selective re-reading only in very difficult syntactic garden-path sentences \citep{paapeReanalysisSelectiveWhen2022,paapeConsciousRereadingConfirmatory2022}. 
In MoTR, participants move their mouse to un-blur a small ``spotlight'' region within a blurry text. By tracking the position of the mouse on the screen, experimenters can compute how long the mouse lingers over each word as a proxy for reading time. 
After each sentence is read, participants select a response from one of the following choices: ``Sentence was OK'', ``I noticed an error'', or ``Not sure''. A breakdown of responses by condition is shown in \Cref{sec:human-responses}. 

\paragraph{Participants} We recruited 200 participants via the online platform Prolific, in accordance with an existing IRB protocol at the authors' institution. 
Participants were self-reported native English speakers residing in the United States, and were paid \$4. The task took approximately 15 minutes. 
Participants provided informed consent before beginning the experiment.
Participants were pseudo-randomly assigned to one of 10 experimental lists, which rotated each item through the 10 variants; each participant thus saw each of the 36 items in exactly 1 condition, and each of the 360 unique item variants was seen by 20 participants. Experimental items were interspersed with 36 filler sentences containing a variety of syntactic constructions and semantic topics, which were seen by all participants.  

\paragraph{Reading measures} The MoTR paradigm results in raw data in the form of many samples of mouse positions on the screen during each trial. Using the post-processing pipeline of \citet{wilcoxMouseTrackingReading2024}, we generate the following continuous reading measures: first fixation duration, gaze duration, go past time, right-bounded reading time, and total duration. We also generate the following binary variables: first pass fixation, first pass regression out, and regression in. Regressions into the \Predicate{} (the final region) are still possible, since the reader may move their mouse beyond the end of the sentence and then back into the sentence. 

\paragraph{Exclusion criteria} We exclude data points based on the following criteria \citep{wilcoxMouseTrackingReading2024}: We exclude data from individual trials if the participant fixated on fewer than 20\% of total words in the sentence. We exclude any words whose gaze duration time was more than 3 standard deviations greater than the mean gaze duration for that word, indicating abnormally long gaze time for that word. We exclude all of a participant's data if they report ``I noticed an error'' for more than 20\% of fillers, which suggests inattentive reading.

\subsection{Computational Modeling}
\label{sec:methods-model}

We employ the model of \citet{clarkModelApproximateIncremental2025} to produce noisy-channel inferences for each sentence in the study \footnote{\url{https://github.com/thomashikaru/noisy_channel_model}; further details are in \Cref{sec:model-details}.}. 
In this model, $\utt$ represents a sequence of observed words, while a set of $K$ weighted particles $\{\xtk{t}{i}\}, i = 1\dots K$, each with weight $\wtk{t}{i}$, represents hypotheses about the model state, e.g. the inferred intended sentence and the sequences of errors that map between the intended sentence and the observed sentence. 
In this model, the prior over strings is derived from the GPT-2 language model, which has been shown to be a strong predictor of human reading times, more so than larger models \citep{shainLargescaleEvidenceLogarithmic2024, ohWhyDoesSurprisal2023}; the model additionally incorporates inferences about possible errors and intended alternatives using Bayesian inference. 

\begin{figure}[htb]
    \centering
    \includegraphics[width=0.9\linewidth]{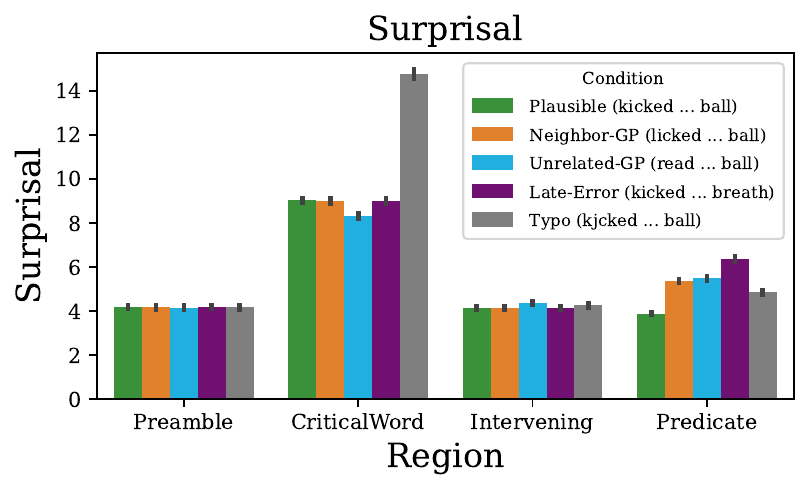}
    \includegraphics[width=0.9\linewidth]{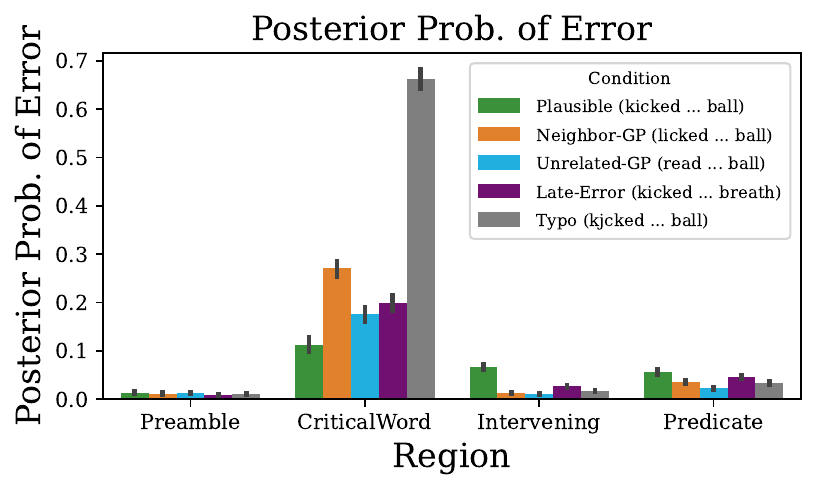}
    \caption{Incremental surprisal and error probability, computed using noisy-channel inference model. Error bars denote 95\% CIs.}
    \label{fig:model-surprisal}
\end{figure}

Two model-based quantities are of particular interest. 
First, we extract incremental surprisal, i.e. the negative log probability of an observed word in context, which is approximated by taking the average particle weight at position $t$:
\begin{align*}
\probP{}(\utt_t \mid \uttab{1}{t-1}) &= \int \probP{}(\utt_t \mid x_t) \probP{}(x_t \mid \uttab{1}{t-1}) dx_t \\ &\approx \frac{1}{K} \sum_{i=1}^K \wtk{t}{i}    
\end{align*}
\Cref{fig:model-surprisal} demonstrates the average per-word surprisal for different regions in the experimental materials, separated by condition. 
The most tightly controlled pair of conditions are \Plausible{} and \NeighborGP, which have identical surprisal values until the \Predicate{} is reached, at which point \NeighborGP{} elicits markedly higher surprisal. The \Typo{} condition is unique in having an immediately obvious error, leading to a large spike in surprisal at the \CriticalWord{}.
While this predictor may explain variance in incremental reading behavior or the probability of a regression being initiated at a particular word, it cannot on its own predict targeted regressions to likely errors earlier in the sentence, in light of new information. 

Second, the model provides an estimate of the posterior distribution over actions at each word in the input utterance, where the \textsc{Normal} action denotes no error, in contrast to the \textsc{Form-Based Substitution}, \textsc{Morphological Substitution} and \textsc{Semantic Substitution} actions (see \Cref{fig:actions-posterior-example} for inferences for an example sentence). 
The posterior probabilities for actions at each word provide an estimate of the probability that, in light of the full utterance, a particular word contained a particular type of error, or no error.  
\Cref{fig:model-surprisal} illustrates the average per-word posterior error probability for different regions, separated by condition. 
Echoing the pattern of surprisal values, the \Typo{} condition has the largest error probability at the \CriticalWord{}. 
Crucially, however, despite the identical surprisal values at the \CriticalWord{} region in \Plausible{} and \NeighborGP{}, the latter has a higher posterior probability of error at the \CriticalWord{} (and is also higher than in \UnrelatedGP{} and \LateError{}). Additional examples of model-generated noisy-channel inferences are provided in \Cref{sec:model-details}.

\subsection{Hypotheses}
\label{sec:methods-hypotheses}

We hypothesize that the results will align with the \textbf{targeted regressions account}: readers will make targeted regressions to the location of likely errors which are promising loci for reanalysis. Under this hypothesis, the probability of regression out from the \Predicate{} region will be higher in all non-\Plausible{} conditions than in the \Plausible{} condition. Furthermore, the \CriticalWord{} in the \NeighborGP{} condition will be associated with a higher probability of regressions in, as well as re-reading time, than in any other condition. We predict that this interaction effect will be greater for \NeighborGP{} than for \UnrelatedGP{}, implying that targeted regressions and re-reading preferentially target errors that are more likely, compared to unrelated word substitutions.  

\subsection{Statistical Analysis}
\label{sec:methods-analysis}

\textbf{Re-reading time: }
We use a Bayesian mixed-effects hurdle-lognormal model for modeling the continuous re-reading (total minus gaze) duration values. Word frequency, word length, incremental surprisal, word position, and part of speech are used to predict the binary value of whether a word will have zero reading time. 
These same predictors, and the additional predictor of a Region $\times$ Condition interaction, are used to predict re-reading duration when it is non-zero. 
Treatment coding is used for categorical predictors; \Plausible{} is used as the reference level for Condition, while \Intervening{} is used as the reference level for Region. In each case, the reference level provides a baseline to which the focal manipulations can be compared. 
Per-participant and per-item random intercepts were included, and per-participant random slopes were included for frequency, length, surprisal, and word position. 
We note that our pre-registration originally did not include re-reading duration as a reading measure, but this was added to capture the specific feature of time spent reading after the first pass. 

\textbf{Regressions in/out: }
We use a Bayesian mixed-effects logistic regression model, with the same predictors and reference levels as above, to predict regressive eye movements, both for a given source and for a given target (\textbf{regressions out} and \textbf{regressions into}). 
Random intercepts and slopes follow the same scheme as for the above models. 

\section{Results}
\label{sec:results}

\subsection{Implausible predicates trigger regressions}

\begin{figure}[htb]
    \centering
    \includegraphics[width=0.9\linewidth]{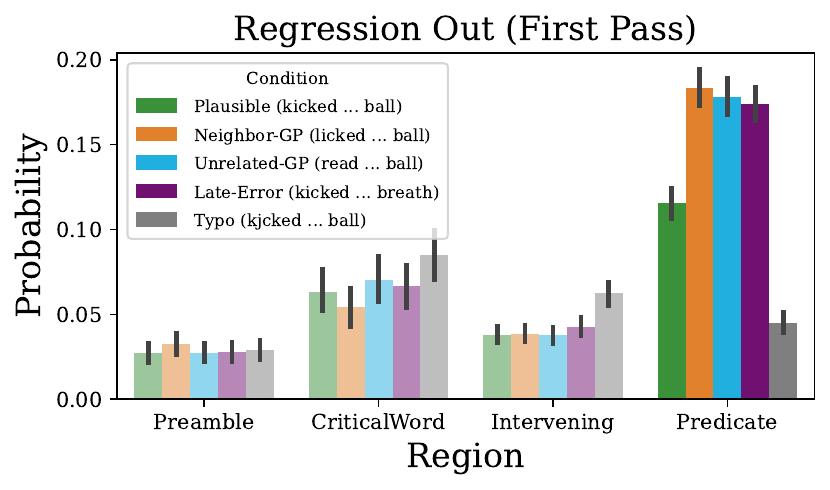}
    \includegraphics[width=0.9\linewidth]{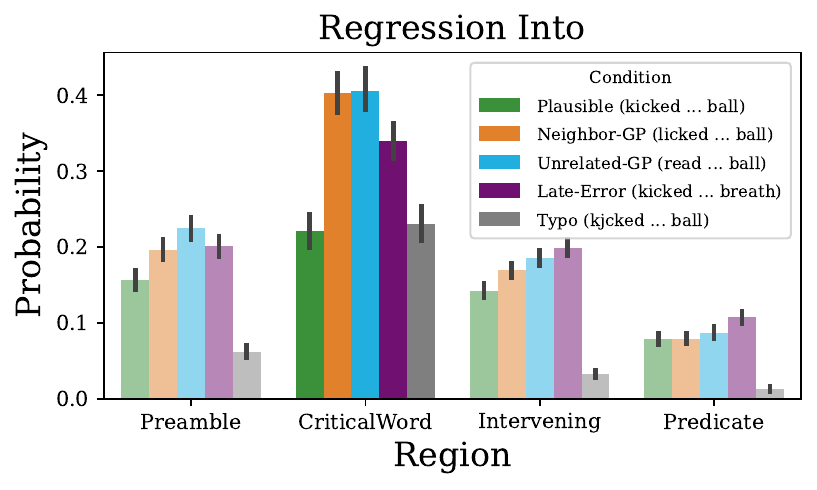}
    \includegraphics[width=0.9\linewidth]{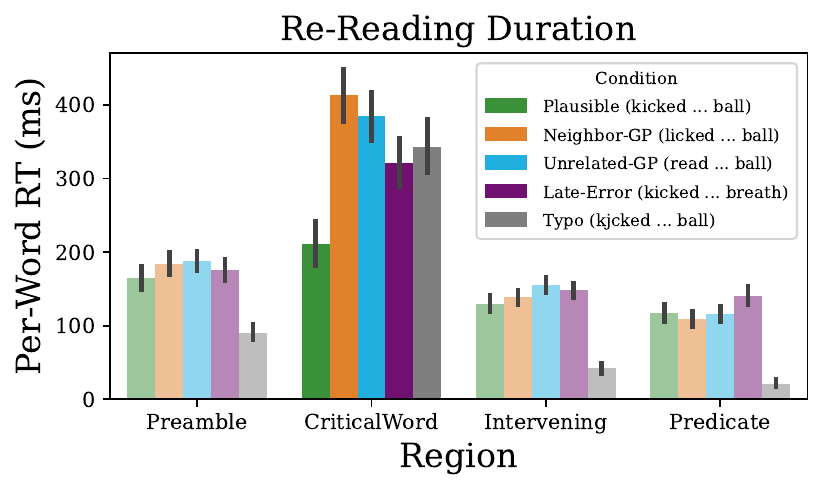}
    \caption{Reading measures as a function of region and condition. Error bars denote 95\% CIs.}
    \label{fig:reading-plots}
\end{figure}

Inspecting the fixed effects of the model predicting regressions out, we observe an interaction between Condition and Region for \NeighborGP{} $\times$ \Predicate{} ($\beta = 0.333$, CrI = [0.196, 0.469]), \UnrelatedGP{} $\times$ \Predicate{} ($\beta = 0.230$, CrI = [0.091, 0.368]), and \LateError{} $\times$ \Predicate{} ($\beta = 0.166$, CrI = [0.028, 0.305]) --- these conditions all had higher probabilities of regressions out at the \Predicate{} than the \Plausible{} (baseline) condition. Meanwhile, the \Typo{} $\times$ \Predicate{} interaction shows a lower probability of regressions out at the \Predicate{}, compared to the \Plausible{} condition ($\beta = -0.450$, CrI = [-0.597, -0.300]), likely because readers tended to already have detected an error earlier in the sentence.  
Full statistical results are presented in \Cref{sec:stats-results}. These results align with the patterns observed in \Cref{fig:reading-plots}.

\subsection{Likely errors attract regressions in}

Inspecting the fixed effects of the model predicting the rate of regressions in, we again observe an interaction between Condition and Region for \NeighborGP{} $\times$ \CriticalWord{} ($\beta = 0.369$, CrI = [0.242, 0.499]) and \UnrelatedGP{} $\times$ \CriticalWord{} ($\beta = 0.279$, CrI = [0.156, 0.406]). 
There was no evidence for an interaction for \LateError{} $\times$ \CriticalWord{} ($\beta = 0.005$, CrI = [-0.118, 0.127]) or \Typo{} $\times$ \CriticalWord{} ($\beta = 0.0$, CrI = [-0.202, 0.189]).
Notably, the largest interaction was observed for the \NeighborGP{} condition, where an orthographic neighbor of the \CriticalWord{} makes for a much more plausible sentence. We see a positive, but numerically smaller interaction in the \UnrelatedGP{} condition, where the \CriticalWord{} is still an ``odd word out'', but there is no obvious mapping from the \CriticalWord{} to a more plausible alternative.

Interactions between Condition and Region were also observed in re-reading time (total minus gaze duration), for \NeighborGP{} $\times$ \CriticalWord{} ($\beta = 0.215$, CrI = [0.125, 0.310]), \Typo{} $\times$ \CriticalWord{} ($\beta = 0.456$, CrI = [0.333, 0.578]), \UnrelatedGP{} $\times$ \CriticalWord{} ($\beta = 0.101$, CrI = [0.007, 0.196]), and \LateError{} $\times$ \CriticalWord{} ($\beta = 0.133$, CrI = [0.037, 0.232]); there was no evidence for other Condition $\times$ Region interactions. 
Full statistical results are presented in \Cref{sec:stats-results}.


\section{Discussion}
\label{sec:discussion}

\subsection{Targeted regressions towards likely errors}

Our results suggest that \textbf{regressions out} are triggered by violations of expectation, found in surprising predicates. 
This is consistent with existing models of incremental processing \citep{levyExpectationbasedSyntacticComprehension2008, haleProbabilisticEarleyParser2001, smithEffectWordPredictability2013}. 
However, what has remained unclear is how readers allocate attention and effort during reanalysis of problematic sentences. 

Turning to the target of regressions, our results indicate that readers are more likely to regress into a word in a sentence which becomes a probable locus of error in light of new information. 
Despite the \Plausible{} and \NeighborGP{} conditions having identical incremental properties up until the \Predicate{}, there are more regressions into and longer re-reading durations on the \CriticalWord{} in \NeighborGP. 
Comparing the \CriticalWord{} to other regions in the \NeighborGP{} condition, we see that the probability of regressions in and re-reading duration are both higher there than at any other region. This rules out the idea that surprising predicates lead to re-reading of the entire sentence (a reasonable alternative explanation), which would result in uniformly higher duration across the whole sentence and more regressions into the \Preamble{} than any other region. 
Comparing the different conditions, we observe that 
\NeighborGP{} and \UnrelatedGP{} have a higher probability of regressions into and higher re-reading duration than the other error conditions, with \NeighborGP{} having a larger interaction estimate than \UnrelatedGP{}. This suggests that reanalysis is sensitive to locations in a sentence where correcting a single error can repair the sentence into something more plausible, especially when this error is a likely one (e.g. an orthographic neighbor).

This targeted behavior could be explained by readers needing time to assess the probability of a language producer making a given error, or because they have uncertainty about the fidelity of their own perceptual input and want to verify or ``double-check'' what the sentence actually said. 
For obvious typos, on the other hand, participants' reading behavior is markedly different --- they tend to linger on it on the first pass, but are less likely to revisit the error. 
The fact that readers show special treatment of typos and orthographic neighbors suggests that readers may employ a mental model of possible errors when interpreting uncertain input \citep{gibsonRationalIntegrationNoisy2013}. 
Our findings are complementary to work finding evidence for targeted regressions that correspond to syntactic reanalysis \citep{timkeyEyeMovementsReveal2025}, showing that an analogous process may be at play in reanalyzing earlier commitments during comprehension as \textit{errors} in light of late-arriving information.

\subsection{Algorithmic accounts of reanalysis}

Existing accounts of eye movements during reading have successfully been applied to different levels of granularity, from the timescale of individual saccades \citep{leggeMrChipsIdealobserver1997,reichleEZReaderModel2003,bicknellOngoingCognitiveProcessing2020}, to high-level information-seeking behavior \citep{grutekekleinEffectSurprisalReading2024}. Yet edge-case behavior, like the reanalysis of earlier words as errors, has not been explicitly modeled in these works.
Meanwhile, works like \citet{levyModelingEffectsMemory2008} model processing of syntactic ambiguity and garden-path sentences with resource-rational approximate inference algorithms, but without a direct linking hypothesis to reanalytical reading behavior when inference fails. 
\citet{clarkModelApproximateIncremental2025} apply a similar class of inference algorithms to the processing of ``noisy'' sentences, where rejuvenation proposals simulate reanalysis of earlier words as errors. This model, however, is under-specified with regard to the ``control flow'' governing the sequential order of inference steps during reading, and supports both unconditional re-reading of a sentence or conditionally triggered reanalysis. In the model, conditional reanalysis instantiates the notion that readers decide whether to reanalyze based on an estimate of the probability that the sentence contains an error, estimated using the surprisal of the current word, normalized by its unigram probability (i.e. ``How surprising is this word \textit{in} context, relative to how surprising it is \textit{out of} context?''). This is closely related to metrics such as \textsc{Slor} \citep{kannSentenceLevelFluencyEvaluation2018} and \textsc{Morcella} \citep{tjuatjaWhatGoesLM2025}. 
Future work can more precisely establish how algorithmic details within approximate probabilistic inference may predict human eye movements, including individual variation.

In this study, predicates in the \NeighborGP{} condition were \textit{not} associated with additional reading duration, compared to minimally different \Plausible{} sentences, but rather with a higher probability of regressions out. This suggests that when re-reading is an option, it is triggered quite readily. 
In contrast, paradigms which do not allow for re-reading (e.g., self-paced reading) have been shown to instead induce slowdowns in spillover regions; notably, this may not be the optimal reading strategy for readers from the perspective of successful comprehension \citep{schotterDontBelieveWhat2014}.
A takeaway for theories of human language processing is that purely incremental processors are insufficient to capture human behavior, and non-optimal from the perspective of handling the uncertainty, ambiguity, and noise present in naturalistic language usage.
Within the approximate probabilistic inference framework for modeling noisy-channel comprehension, reanalysis is strongly motivated as a resource-rational, efficient strategy: to guarantee recovery from errors in purely incremental processing would require a massive amount of parallel computation and memory to encode all possible alternatives of a linguistic unit, just in case this alternative is later rendered likely by new information. 
In contrast, allowing the model to execute rejuvenation moves only when an ``error signal'' is detected would allow processing to proceed with far fewer cognitive resources. 
Since errors are generally infrequent compared to non-errors, but are crucial to correct when they occur, a flexible strategy capable of deploying reanalysis when needed is an efficient solution to the possibly-noisy language comprehension problem. 
More broadly, this framework has implications for the processing of syntactic garden-path sentences as well: many garden-path sentences are amenable to alternative interpretations in terms of missing or substituted words rather than syntax (e.g. positions $A$ and $B$ in \textit{The bird [that]$_A$ perched on the branch [and]$_B$ sang sweetly}). 

\subsection{Syntactic dependencies and PMI}

Past work has suggested that reading regressions align with both the latent syntactic dependency structure of a sentence \citep{lopopoloDependencyParsingYour2019} as well as with the PMI between pairs of words in a sentence \citep{wilcoxInformationtheoreticAnalysisTargeted2024}; dependency relations are also known to be associated with high PMI \citep{futrellInformationtheoreticLocalityProperties2019a}.
Our materials, which manipulate the plausibility of a sentence within each item while preserving the same syntactic structure, provide a way to disentangle the influence of dependency structure and PMI. 
\Cref{sec:appendix-pmi} reports a systematic comparison of the PMI between \CriticalWord{} and \Predicate{} across conditions; as would be expected, \Plausible{} items have significantly higher PMI between these two regions than \NeighborGP{} items. 
The dependency structure of an example garden-path sentence is shown below, with a bolded arrow indicating a dependency arc between \Predicate{} and \CriticalWord{}:

\begin{center}
\footnotesize
\begin{dependency}[arc edge, arc angle=80, text only label, label style={above},edge style={gray}]
\begin{deptext}[column sep=.1cm]
The \& boy \& licked \& the \& big \& round \& \textbf{ball} \& into \& the \& net.\\
\& \& \& \& \& \& \textbf{\faEye} \& \& \&\\
\end{deptext}
\deproot{3}{root}
\depedge{2}{1}{det}
\depedge{3}{2}{nsubj}
\depedge[edge style={black, ultra thick}]{3}{7}{obj}
\depedge{3}{10}{obl}
\depedge{7}{4}{det}
\depedge[label style={below}]{7}{5}{amod}
\depedge[label style={below}]{7}{6}{amod}
\depedge{10}{8}{case}
\depedge[label style={below}]{10}{9}{det}
\end{dependency}
\end{center}

Our results show that previously read words that are in a given dependency relation are \textit{more} likely to be regressed to when source and target are low-PMI (\textit{licked, ball}) than when they are high-PMI (\textit{licked, lollipop}). 
This provides evidence for reanalytical regressions (as opposed to reactivation regressions), which was inconclusive in the naturalistic reading time corpus study of \citet{wilcoxInformationtheoreticAnalysisTargeted2024}. 
In naturalistic corpora, high PMI might predict regressions better than low or negative PMI in aggregate, while direct experimental manipulation shows that violations of expectation can drive regressions along dependency arcs when raw PMI would not predict this. 

One interpretation of this pattern is that readers maintain a hierarchical mental representation corresponding to dependencies between words, independently of the statistical co-occurrence patterns of words; thus a reader knows that in the example above, the word \textit{licked} is the head of \textit{ball}, and makes a regression from dependent to head to attempt to resolve the violation of expectation.
Another interpretation is that the reader may have access to other predictions about the upcoming word, e.g. in the sentence above, a reader may may predict \textit{lollipop} to follow \textit{licked the big round...}; upon seeing the unexpected word \textit{ball}, the reader regresses to the word in the sentence that has high PMI with the predicted but unseen word. The latter account preserves the usefulness of PMI as a proxy for dependency relations, but requires considering the PMI between \textit{predicted} words and earlier words, rather than only that of observed words. 


\section{Conclusion}
\label{sec:conclusion}

In this study, we collected a large dataset comprising mouse-tracking data from 200 participants on 360 unique experimental sentences. 
Using these data, we present empirical results showing re-reading behavior that is targeted to the locus of plausible errors. 
This behavior is qualitatively consistent with the condition-level differences in posterior error probability of an algorithmic model of noisy-channel processing, which performs explicit reanalysis to update beliefs about earlier commitments in light of new information. 
Building on past work on information-theoretic and rational-inference approaches to language comprehension, we provide an integrated explanation of regressive readings during reanalysis, broadly construed --- not just syntactic reanalysis, but reanalysis of earlier materials as possible errors.
These results demonstrate the flexibility of readers in forming interpretations of noisy language input, and the fine-grained processes underlying robust comprehension during reading.

\section*{Limitations}

One limitation of this study is the divergence from a truly naturalistic reading environment. 
Readers may be adopting strategies to answer the questions with minimal effort (for example, skipping the rest of the sentence as soon as a typo is read). Although this is a limitation, it also is a source of insights: in comparison to the \Typo{} conditions, we can be relatively confident that participants are actually reading through the other conditions attentively, and our data show  a glimpse into the timecourse of inferences relating to whether a sentence contains an error or not. 
In the case of non-word errors, this is an easy task, but in sentences with semantic or lexical anomalies, readers make reading regressions to interrogate earlier parts of a sentence before making a final decision. 
 
Individual variation between readers, who may adopt qualitatively different strategies for handling anomalous inputs, is likewise not fully investigated in this study.
Data like those collected in this study, which show fine-grained reading behavior for a range of plausible and implausible sentences, provide the potential for new insights towards this question. The pattern of regressions, including the conditions which trigger a regression from a source to a target word, can inform the inference control flow in resource-rational models. 

Additionally, this study's exclusive focus on English limits the generality of its findings, and future work can consider typologically diverse languages to gain insight into reanalysis strategies for reading in languages with significantly different writing systems, word order, or morphological complexity. 


\bibliography{custom}

@article{aaronsonPerformanceTheoriesSentence1976,
  title = {Performance Theories for Sentence Coding: {{Some}} Quantitative Evidence},
  shorttitle = {Performance Theories for Sentence Coding},
  author = {Aaronson, Doris and Scarborough, Hollis S.},
  year = 1976,
  journal = {Journal of Experimental Psychology: Human Perception and Performance},
  volume = {2},
  number = {1},
  pages = {56--70},
  publisher = {American Psychological Association},
  address = {US},
  issn = {1939-1277},
  doi = {10.1037/0096-1523.2.1.56}
}

@article{bicknellOngoingCognitiveProcessing2020,
  title = {Ongoing {{Cognitive Processing Influences Precise Eye-Movement Targets}} in {{Reading}}},
  author = {Bicknell, Klinton and Levy, Roger and Rayner, Keith},
  year = 2020,
  journal = {Psychological Science},
  volume = {31},
  number = {4},
  pages = {351--362},
  issn = {0956-7976},
  doi = {10.1177/0956797620901766},
  pmcid = {PMC7436780},
  pmid = {32105193}
}

@article{christiansonRetracingGardenpathNonselective2024,
  title = {Retracing the Garden-Path: {{Nonselective}} Rereading and No Reanalysis},
  shorttitle = {Retracing the Garden-Path},
  author = {Christianson, Kiel and Dempsey, Jack and Tsiola, Anna and Deshaies, Sarah-Elizabeth M. and Kim, Nayoung},
  year = 2024,
  journal = {Journal of Memory and Language},
  volume = {137},
  pages = {104515},
  issn = {0749-596X},
  doi = {10.1016/j.jml.2024.104515}
}

@article{clarkModelApproximateIncremental2025,
  title = {A {{Model}} of {{Approximate}} and {{Incremental Noisy-Channel Language Processing}}},
  author = {Clark, Thomas and Vigly, Jacob Hoover and Gibson, Edward and Levy, Roger},
  year = 2025,
  journal = {Proceedings of the Annual Meeting of the Cognitive Science Society},
  volume = {47},
  number = {0}
}

@inproceedings{clarkResourceRationalNoisyChannelLanguage2025,
  title = {Resource-{{Rational Noisy-Channel Language Processing}}: {{Testing}} the {{Effect}} of {{Algorithmic Constraints}} on {{Inferences}}},
  shorttitle = {Resource-{{Rational Noisy-Channel Language Processing}}},
  booktitle = {Proceedings of the 2025 {{Conference}} on {{Empirical Methods}} in {{Natural Language Processing}}},
  author = {Clark, Thomas Hikaru and Vigly, Jacob Hoover and Gibson, Edward and Levy, Roger P.},
  editor = {Christodoulopoulos, Christos and Chakraborty, Tanmoy and Rose, Carolyn and Peng, Violet},
  year = 2025,
  pages = {23659--23672},
  publisher = {Association for Computational Linguistics},
  address = {Suzhou, China}
}

@incollection{doucetIntroductionSequentialMonte2001,
  title = {An {{Introduction}} to {{Sequential Monte Carlo Methods}}},
  booktitle = {Sequential {{Monte Carlo Methods}} in {{Practice}}},
  author = {Doucet, Arnaud and {de Freitas}, Nando and Gordon, Neil},
  editor = {Doucet, Arnaud and {de Freitas}, Nando and Gordon, Neil},
  year = 2001,
  series = {Statistics for {{Engineering}} and {{Information Science}}},
  pages = {3--14},
  publisher = {Springer},
  address = {New York, NY},
  doi = {10.1007/978-1-4757-3437-9_1}
}

@article{forsterMazeTaskMeasuring2009,
  title = {The Maze Task: {{Measuring}} Forced Incremental Sentence Processing Time},
  shorttitle = {The Maze Task},
  author = {Forster, Kenneth I. and Guerrera, Christine and Elliot, Lisa},
  year = 2009,
  journal = {Behavior Research Methods},
  volume = {41},
  number = {1},
  pages = {163--171},
  issn = {1554-3528},
  doi = {10.3758/BRM.41.1.163}
}

@article{frazierMakingCorrectingErrors1982,
  title = {Making and Correcting Errors during Sentence Comprehension: {{Eye}} Movements in the Analysis of Structurally Ambiguous Sentences},
  shorttitle = {Making and Correcting Errors during Sentence Comprehension},
  author = {Frazier, Lyn and Rayner, Keith},
  year = 1982,
  journal = {Cognitive Psychology},
  volume = {14},
  number = {2},
  pages = {178--210},
  issn = {0010-0285},
  doi = {10.1016/0010-0285(82)90008-1}
}

@inproceedings{futrellInformationtheoreticLocalityProperties2019a,
  title = {Information-Theoretic Locality Properties of Natural Language},
  booktitle = {Proceedings of the {{First Workshop}} on {{Quantitative Syntax}} ({{Quasy}}, {{SyntaxFest}} 2019)},
  author = {Futrell, Richard},
  editor = {Chen, Xinying and {Ferrer-i-Cancho}, Ramon},
  year = 2019,
  pages = {2--15},
  publisher = {Association for Computational Linguistics},
  address = {Paris, France},
  doi = {10.18653/v1/W19-7902}
}

@article{futrellNaturalStoriesCorpus2021,
  title = {The {{Natural Stories}} Corpus: A Reading-Time Corpus of {{English}} Texts Containing Rare Syntactic Constructions},
  shorttitle = {The {{Natural Stories}} Corpus},
  author = {Futrell, Richard and Gibson, Edward and Tily, Harry J. and Blank, Idan and Vishnevetsky, Anastasia and Piantadosi, Steven T. and Fedorenko, Evelina},
  year = 2021,
  journal = {Language Resources and Evaluation},
  volume = {55},
  number = {1},
  pages = {63--77},
  issn = {1574-0218},
  doi = {10.1007/s10579-020-09503-7}
}

@article{gibsonRationalIntegrationNoisy2013,
  title = {Rational Integration of Noisy Evidence and Prior Semantic Expectations in Sentence Interpretation},
  author = {Gibson, Edward and Bergen, Leon and Piantadosi, Steven T.},
  year = 2013,
  journal = {Proceedings of the National Academy of Sciences},
  volume = {110},
  number = {20},
  pages = {8051--8056},
  publisher = {Proceedings of the National Academy of Sciences},
  doi = {10.1073/pnas.1216438110}
}

@inproceedings{haleProbabilisticEarleyParser2001,
  title = {A Probabilistic {{Earley}} Parser as a Psycholinguistic Model},
  booktitle = {Second {{Meeting}} of the {{North American Chapter}} of the {{Association}} for {{Computational Linguistics}}},
  author = {Hale, John},
  year = 2001
}

@inproceedings{hooverLinguisticDependenciesStatistical2021a,
  title = {Linguistic {{Dependencies}} and {{Statistical Dependence}}},
  booktitle = {Proceedings of the 2021 {{Conference}} on {{Empirical Methods}} in {{Natural Language Processing}}},
  author = {Hoover, Jacob Louis and Du, Wenyu and Sordoni, Alessandro and O'Donnell, Timothy J.},
  editor = {Moens, Marie-Francine and Huang, Xuanjing and Specia, Lucia and Yih, Scott Wen-tau},
  year = 2021,
  pages = {2941--2963},
  publisher = {Association for Computational Linguistics},
  address = {Online and Punta Cana, Dominican Republic},
  doi = {10.18653/v1/2021.emnlp-main.234}
}

@article{leggeMrChipsIdealobserver1997,
  title = {Mr. {{Chips}}: {{An}} Ideal-Observer Model of Reading},
  shorttitle = {Mr. {{Chips}}},
  author = {Legge, Gordon E. and Klitz, Timothy S. and Tjan, Bosco S.},
  year = 1997,
  journal = {Psychological Review},
  volume = {104},
  number = {3},
  pages = {524--553},
  publisher = {American Psychological Association},
  address = {US},
  issn = {1939-1471},
  doi = {10.1037/0033-295X.104.3.524}
}

@article{levyExpectationbasedSyntacticComprehension2008,
  title = {Expectation-Based Syntactic Comprehension},
  author = {Levy, Roger},
  year = 2008,
  journal = {Cognition},
  volume = {106},
  pages = {1126--1177},
  publisher = {Elsevier Science},
  address = {Netherlands},
  issn = {1873-7838},
  doi = {10.1016/j.cognition.2007.05.006}
}

@inproceedings{levyModelingEffectsMemory2008,
  title = {Modeling the Effects of Memory on Human Online Sentence Processing with Particle Filters},
  booktitle = {Advances in {{Neural Information Processing Systems}}},
  author = {Levy, Roger and Reali, Florencia and Griffiths, Thomas},
  year = 2008,
  volume = {21},
  publisher = {Curran Associates, Inc.}
}

@inproceedings{levyNoisyChannelModelHuman2008,
  title = {A {{Noisy-Channel Model}} of {{Human Sentence Comprehension}} under {{Uncertain Input}}},
  booktitle = {Proceedings of the 2008 {{Conference}} on {{Empirical Methods}} in {{Natural Language Processing}}},
  author = {Levy, Roger},
  year = 2008,
  pages = {234--243},
  publisher = {Association for Computational Linguistics},
  address = {Honolulu, Hawaii}
}

@misc{lewSequentialMonteCarlo2023,
  title = {Sequential {{Monte Carlo Steering}} of {{Large Language Models}} Using {{Probabilistic Programs}}},
  author = {Lew, Alexander K. and {Zhi-Xuan}, Tan and Grand, Gabriel and Mansinghka, Vikash K.},
  year = 2023,
  number = {arXiv:2306.03081},
  eprint = {2306.03081},
  primaryclass = {cs},
  publisher = {arXiv},
  doi = {10.48550/arXiv.2306.03081},
  archiveprefix = {arXiv}
}

@article{liHeuristicInterpretationRational2023,
  title = {Heuristic Interpretation as Rational Inference: {{A}} Computational Model of the {{N400}} and {{P600}} in Language Processing},
  shorttitle = {Heuristic Interpretation as Rational Inference},
  author = {Li, Jiaxuan and Ettinger, Allyson},
  year = 2023,
  journal = {Cognition},
  volume = {233},
  pages = {105359},
  issn = {0010-0277},
  doi = {10.1016/j.cognition.2022.105359}
}

@misc{liInformationtheoreticModelShallow2024,
  title = {An Information-Theoretic Model of Shallow and Deep Language Comprehension},
  author = {Li, Jiaxuan and Futrell, Richard},
  year = 2024,
  number = {arXiv:2405.08223},
  eprint = {2405.08223},
  primaryclass = {cs, math},
  publisher = {arXiv},
  archiveprefix = {arXiv}
}

@inproceedings{lopopoloDependencyParsingYour2019,
  title = {Dependency {{Parsing}} with Your {{Eyes}}: {{Dependency Structure Predicts Eye Regressions During Reading}}},
  shorttitle = {Dependency {{Parsing}} with Your {{Eyes}}},
  booktitle = {Proceedings of the {{Workshop}} on {{Cognitive Modeling}} and {{Computational Linguistics}}},
  author = {Lopopolo, Alessandro and Frank, Stefan L. and {van den Bosch}, Antal and Willems, Roel},
  editor = {Chersoni, Emmanuele and Jacobs, Cassandra and Lenci, Alessandro and Linzen, Tal and Pr{\'e}vot, Laurent and Santus, Enrico},
  year = 2019,
  pages = {77--85},
  publisher = {Association for Computational Linguistics},
  address = {Minneapolis, Minnesota},
  doi = {10.18653/v1/W19-2909}
}

@article{mitchellAccountingRegressiveEyemovements2008,
  title = {Accounting for Regressive Eye-Movements in Models of Sentence Processing: {{A}} Reappraisal of the {{Selective Reanalysis}} Hypothesis},
  shorttitle = {Accounting for Regressive Eye-Movements in Models of Sentence Processing},
  author = {Mitchell, Don C. and Shen, Xingjia and Green, Matthew J. and Hodgson, Timothy L.},
  year = 2008,
  journal = {Journal of Memory and Language},
  volume = {59},
  number = {3},
  pages = {266--293},
  issn = {0749-596X},
  doi = {10.1016/j.jml.2008.06.002}
}

@article{mitchellEffectsContextContent1978,
  title = {The {{Effects}} of {{Context}} and {{Content}} on {{Immediate Processing}} in {{Reading}}},
  author = {Mitchell, D. C. and Green, D. W.},
  year = 1978,
  journal = {Quarterly Journal of Experimental Psychology},
  volume = {30},
  number = {4},
  pages = {609--636},
  publisher = {SAGE Publications},
  issn = {0033-555X},
  doi = {10.1080/14640747808400689}
}

@misc{naessethElementsSequentialMonte2024,
  title = {Elements of {{Sequential Monte Carlo}}},
  author = {Naesseth, Christian A. and Lindsten, Fredrik and Sch{\"o}n, Thomas B.},
  year = 2024,
  number = {arXiv:1903.04797},
  eprint = {1903.04797},
  primaryclass = {stat},
  publisher = {arXiv},
  doi = {10.48550/arXiv.1903.04797},
  archiveprefix = {arXiv}
}

@article{paapeDoesLocalCoherence2021,
  title = {Does {{Local Coherence Lead}} to {{Targeted Regressions}} and {{Illusions}} of {{Grammaticality}}?},
  author = {Paape, Dario and Vasishth, Shravan and Engbert, Ralf},
  year = 2021,
  journal = {Open Mind},
  volume = {5},
  pages = {42--58},
  issn = {2470-2986},
  doi = {10.1162/opmi_a_00041}
}

@article{paapeEstimatingTrueCost2022,
  title = {Estimating the {{True Cost}} of {{Garden Pathing}}: {{A Computational Model}} of {{Latent Cognitive Processes}}},
  shorttitle = {Estimating the {{True Cost}} of {{Garden Pathing}}},
  author = {Paape, Dario and Vasishth, Shravan},
  year = 2022,
  journal = {Cognitive Science},
  volume = {46},
  number = {8},
  pages = {e13186},
  issn = {1551-6709},
  doi = {10.1111/cogs.13186},
  copyright = {\copyright{} 2022 The Authors. Cognitive Science published by Wiley Periodicals LLC on behalf of Cognitive Science Society (CSS).}
}

@article{reichleEZReaderModel2003,
  title = {The {{E-Z Reader}} Model of Eye-Movement Control in Reading: {{Comparisons}} to Other Models},
  shorttitle = {The {{E-Z Reader}} Model of Eye-Movement Control in Reading},
  author = {Reichle, Erik D. and Rayner, Keith and Pollatsek, Alexander},
  year = 2003,
  journal = {Behavioral and Brain Sciences},
  volume = {26},
  number = {4},
  pages = {445--476},
  issn = {0140-525X, 1469-1825},
  doi = {10.1017/S0140525X03000104},
  copyright = {https://www.cambridge.org/core/terms}
}

@article{schotterDontBelieveWhat2014,
  title = {Don't {{Believe What You Read}} ({{Only Once}}): {{Comprehension Is Supported}} by {{Regressions During Reading}}},
  shorttitle = {Don't {{Believe What You Read}} ({{Only Once}})},
  author = {Schotter, Elizabeth R. and Tran, Randy and Rayner, Keith},
  year = 2014,
  journal = {Psychological Science},
  volume = {25},
  number = {6},
  pages = {1218--1226},
  publisher = {SAGE Publications Inc},
  issn = {0956-7976},
  doi = {10.1177/0956797614531148}
}

@article{shainLargescaleEvidenceLogarithmic2024,
  title = {Large-Scale Evidence for Logarithmic Effects of Word Predictability on Reading Time},
  author = {Shain, Cory and Meister, Clara and Pimentel, Tiago and Cotterell, Ryan and Levy, Roger},
  year = 2024,
  journal = {Proceedings of the National Academy of Sciences},
  volume = {121},
  number = {10},
  pages = {e2307876121},
  publisher = {Proceedings of the National Academy of Sciences},
  doi = {10.1073/pnas.2307876121}
}

@article{smithEffectWordPredictability2013,
  title = {The Effect of Word Predictability on Reading Time Is Logarithmic},
  author = {Smith, Nathaniel J. and Levy, Roger},
  year = 2013,
  journal = {Cognition},
  volume = {128},
  number = {3},
  pages = {302--319},
  issn = {0010-0277},
  doi = {10.1016/j.cognition.2013.02.013}
}

@misc{tjuatjaWhatGoesLM2025,
  title = {What {{Goes Into}} a {{LM Acceptability Judgment}}? {{Rethinking}} the {{Impact}} of {{Frequency}} and {{Length}}},
  shorttitle = {What {{Goes Into}} a {{LM Acceptability Judgment}}?},
  author = {Tjuatja, Lindia and Neubig, Graham and Linzen, Tal and Hao, Sophie},
  year = 2025,
  number = {arXiv:2411.02528},
  eprint = {2411.02528},
  primaryclass = {cs},
  publisher = {arXiv},
  doi = {10.48550/arXiv.2411.02528},
  archiveprefix = {arXiv}
}

@article{wilcoxInformationtheoreticAnalysisTargeted2024,
  title = {An Information-Theoretic Analysis of Targeted Regressions during Reading},
  author = {Wilcox, Ethan Gotlieb and Pimentel, Tiago and Meister, Clara and Cotterell, Ryan},
  year = 2024,
  journal = {Cognition},
  volume = {249},
  pages = {105765},
  issn = {0010-0277},
  doi = {10.1016/j.cognition.2024.105765}
}

@article{wilcoxMouseTrackingReading2024,
  title = {Mouse {{Tracking}} for {{Reading}} ({{MoTR}}): {{A}} New Naturalistic Incremental Processing Measurement Tool},
  shorttitle = {Mouse {{Tracking}} for {{Reading}} ({{MoTR}})},
  author = {Wilcox, Ethan Gotlieb and Ding, Cui and Sachan, Mrinmaya and J{\"a}ger, Lena Ann},
  year = 2024,
  journal = {Journal of Memory and Language},
  volume = {138},
  pages = {104534},
  issn = {0749-596X},
  doi = {10.1016/j.jml.2024.104534}
}

@article{ohWhyDoesSurprisal2023,
    title = {Why does surprisal from larger transformer-based language models provide a poorer fit to human reading times?},
    volume = {11},
    url = {https://aclanthology.org/2023.tacl-1.20},
    doi = {10.1162/tacl_a_00548},
    urldate = {2024-10-22},
    journal = {Transactions of the Association for Computational Linguistics},
    author = {Oh, Byung-Doh and Schuler, William},
    year = {2023},
    note = {Place: Cambridge, MA
Publisher: MIT Press},
    pages = {336--350},
}

@article{ryskinERPIndexRealtime2021,
    title = {An {ERP} index of real-time error correction within a noisy-channel framework of human communication},
    volume = {158},
    issn = {1873-3514},
    doi = {10.1016/j.neuropsychologia.2021.107855},
    language = {eng},
    journal = {Neuropsychologia},
    author = {Ryskin, Rachel and Stearns, Laura and Bergen, Leon and Eddy, Marianna and Fedorenko, Evelina and Gibson, Edward},
    month = jul,
    year = {2021},
    pages = {107855},
}

@article{raynerEyeMovementsReading1998,
  title = {Eye Movements in Reading and Information Processing: 20 Years of Research},
  shorttitle = {Eye Movements in Reading and Information Processing},
  author = {Rayner, Keith},
  year = 1998,
  journal = {Psychological Bulletin},
  volume = {124},
  number = {3},
  pages = {372--422},
  publisher = {American Psychological Association},
  address = {US},
  issn = {1939-1455},
  doi = {10.1037/0033-2909.124.3.372}
}

@incollection{beverCognitiveBasisLinguistic1970,
    title = {The {Cognitive} {Basis} for {Linguistic} {Structures}},
    isbn = {978-0-19-967713-9},
    doi = {10.1093/acprof:oso/9780199677139.003.0001},
    author = {Bever, Thomas},
    month = jan,
    year = {1970},
    pages = {279--352},
}

@article{frazierComprehendingSentencesSyntactic1979,
    title = {On {Comprehending} {Sentences}: {Syntactic} {Parsing} {Strategies}},
    shorttitle = {On {Comprehending} {Sentences}},
    journal = {ETD Collection for University of Connecticut},
    author = {Frazier, Lyn},
    month = jan,
    year = {1979},
}

@article{paapeReanalysisSelectiveWhen2022,
  title = {Is Reanalysis Selective When Regressions Are Consciously Controlled?},
  author = {Paape, Dario and Vasishth, Shravan},
  year = 2022,
  journal = {Glossa Psycholinguistics},
  volume = {1},
  number = {1},
  issn = {2767-0279},
  doi = {10.5070/G601139},
  copyright = {Copyright: \copyright{} 2021 The Author(s). This is an open-access article distributed under the terms of the Creative Commons Attribution 4.0 International License (CC-BY 4.0), which permits unrestricted use, distribution, and reproduction in any medium, provided the original author and source are credited. See http://creativecommons.org/licenses/by/4.0/.}
}

@article{paapeConsciousRereadingConfirmatory2022,
  title = {Conscious Rereading Is Confirmatory: {{Evidence}} from Bidirectional Self-Paced Reading},
  shorttitle = {Conscious Rereading Is Confirmatory},
  author = {Paape, Dario and Vasishth, Shravan},
  year = 2022,
  journal = {Glossa Psycholinguistics},
  volume = {1},
  number = {1},
  issn = {2767-0279},
  doi = {10.5070/G6011182},
  copyright = {Copyright: \copyright{} 2022 The Author(s). This is an open-access article distributed under the terms of the Creative Commons Attribution 4.0 International License (CC-BY 4.0), which permits unrestricted use, distribution, and reproduction in any medium, provided the original author and source are credited. See http://creativecommons.org/licenses/by/4.0/.}
}

@misc{timkeyEyeMovementsReveal2025,
  title = {Eye Movements Reveal a Dissociation between Prediction and Structural Processing in Language Comprehension},
  author = {Timkey, William and Huang, Kuan-Jung and Oh, Byung-Doh and Prasad, Grusha and Arehalli, Suhas and Linzen, Tal and Dillon, Brian},
  year = 2025,
  number = {eq2ra\_v1},
  publisher = {PsyArXiv},
  doi = {10.31234/osf.io/eq2ra_v1}
}

@inproceedings{grutekekleinEffectSurprisalReading2024,
  title = {The {{Effect}} of {{Surprisal}} on {{Reading Times}} in {{Information Seeking}} and {{Repeated Reading}}},
  booktitle = {Proceedings of the 28th {{Conference}} on {{Computational Natural Language Learning}}},
  author = {Gruteke Klein, Keren and Meiri, Yoav and Shubi, Omer and Berzak, Yevgeni},
  editor = {Barak, Libby and Alikhani, Malihe},
  year = 2024,
  pages = {219--230},
  publisher = {Association for Computational Linguistics},
  address = {Miami, FL, USA},
  doi = {10.18653/v1/2024.conll-1.17}
}

@inproceedings{kannSentenceLevelFluencyEvaluation2018,
  title = {Sentence-{{Level Fluency Evaluation}}: {{References Help}}, {{But Can Be Spared}}!},
  shorttitle = {Sentence-{{Level Fluency Evaluation}}},
  booktitle = {Proceedings of the 22nd {{Conference}} on {{Computational Natural Language Learning}}},
  author = {Kann, Katharina and Rothe, Sascha and Filippova, Katja},
  editor = {Korhonen, Anna and Titov, Ivan},
  year = 2018,
  pages = {313--323},
  publisher = {Association for Computational Linguistics},
  address = {Brussels, Belgium},
  doi = {10.18653/v1/K18-1031}
}

@article{brysbaertMovingKuceraFrancis2009,
    title = {Moving beyond {Kučera} and {Francis}: {A} critical evaluation of current word frequency norms and the introduction of a new and improved word frequency measure for {American} {English}},
    volume = {41},
    issn = {1554-351X, 1554-3528},
    shorttitle = {Moving beyond kučera and francis},
    url = {http://link.springer.com/10.3758/BRM.41.4.977},
    doi = {10.3758/BRM.41.4.977},
    language = {en},
    number = {4},
    urldate = {2024-01-30},
    journal = {Behavior Research Methods},
    author = {Brysbaert, Marc and New, Boris},
    month = nov,
    year = {2009},
    pages = {977--990},
}

@misc{liuRoBERTaRobustlyOptimized2019a,
    title = {{RoBERTa}: {A} {Robustly} {Optimized} {BERT} {Pretraining} {Approach}},
    shorttitle = {{RoBERTa}},
    url = {http://arxiv.org/abs/1907.11692},
    doi = {10.48550/arXiv.1907.11692},
    urldate = {2026-05-16},
    publisher = {arXiv},
    author = {Liu, Yinhan and Ott, Myle and Goyal, Naman and Du, Jingfei and Joshi, Mandar and Chen, Danqi and Levy, Omer and Lewis, Mike and Zettlemoyer, Luke and Stoyanov, Veselin},
    month = jul,
    year = {2019},
    note = {arXiv:1907.11692 [cs.CL]},
}


\appendix

\FloatBarrier

\section{Readers' sentence ratings differentiate categories of possible errors}
\label{sec:human-responses}

\begin{figure}[htb]
    \centering
    \includegraphics[width=0.9\linewidth]{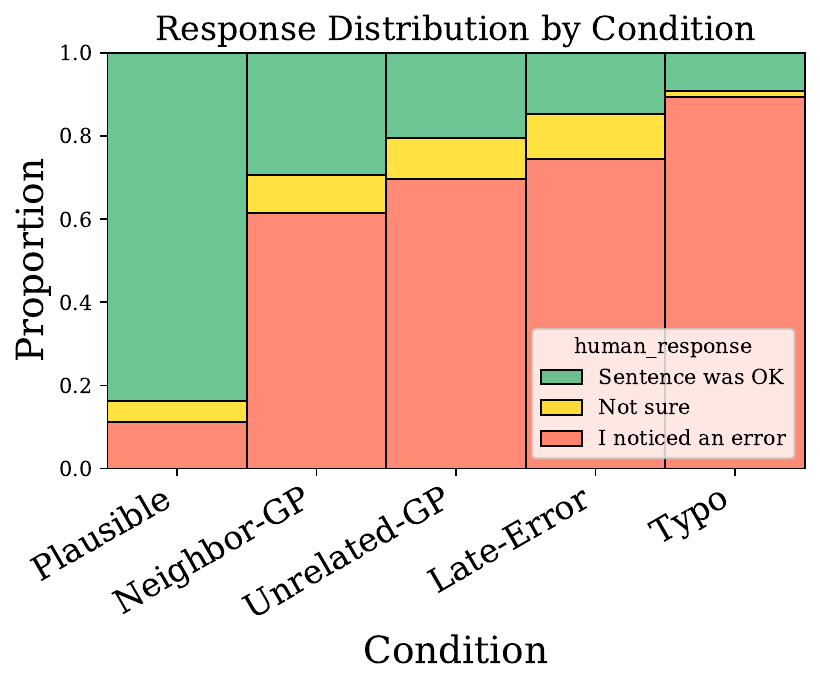}
    \caption{Participant response by condition.}
    \label{fig:response-by-cond}
\end{figure}

After reading each sentence, participants were asked to select a response regarding whether the sentence contained an error, with the options of ``Sentence was OK'', ``I noticed an error'', and ``Not sure''. 
Participants were most likely to report an error for the \Typo{} condition, followed by \LateError{}, then \UnrelatedGP{}, then \NeighborGP, then Plausible (\Cref{fig:response-by-cond}).
This provides evidence that participants were performing the task properly.
Additionally, we observe that the \Typo{} condition led to very little uncertainty in judgments and very low rates of acceptability. 

\section{Additional Experimental Materials}
\label{sec:extra-examples}

\tablecaption{Experimental sentences showing \CriticalWord{} and \Predicate{} manipulations. \NeighborGP{} sentences result when \Predicate{} and \CriticalWord{} are mismatched, while Plausible sentences result from matched pairs. Other conditions are not shown.}
\label{tab:sentences}
\tablefirsthead{
\hline
\textbf{Sentence} \\
\hline
}
\tablehead{
\hline
\textbf{Sentence} \\
\hline
}
\begin{supertabular}{p{0.97\columnwidth}}
The boy \textbf{kicked}/\textit{licked} the big round \textbf{ball into the net}/\textit{lollipop with delight}. \\
His \textbf{band}/\textit{hand} was apparently no longer \textbf{performing live}/\textit{bandaged tightly}. \\
In the \textbf{base}/\textit{vase} were a wide range of \textbf{soldiers from all over}/\textit{flowers from my mom}. \\
She got a lovely \textbf{fan}/\textit{tan} while spending some time in the \textbf{gift shop this morning}/\textit{sun over the holidays}. \\
The \textbf{car}/\textit{care} that was provided by the \textbf{dealer was excellent}/\textit{doctor was excellent}. \\
The doctor \textbf{wired}/\textit{wiped} the heavily sedated patient's \textbf{jaws shut during the surgery}/\textit{blood off during the surgery}. \\
Can you please \textbf{dip}/\textit{zip} that big black \textbf{brush into the paint}/\textit{bag up}? \\
The student \textbf{scored}/\textit{scared} a very difficult \textbf{goal with a header}/\textit{teacher with a prank}. \\
The \textbf{tires}/\textit{fires} were quickly and efficiently \textbf{pumped up by the mechanic}/\textit{extinguished by the firefighters}. \\
The \textbf{bears}/\textit{beers} were last seen \textbf{roaming in the forest}/\textit{chilling in the freezer}. \\
The driver \textbf{bumped}/\textit{pumped} the bright red \textbf{traffic cone with his car}/\textit{tank full of gas}. \\
Why did you \textbf{shave}/\textit{shove} your really nice \textbf{beard for no reason}/\textit{teammate for no reason}? \\
The \textbf{checks}/\textit{chicks} were unexpectedly \textbf{cashed on Monday}/\textit{hatched on Monday}. \\
I watched as a \textbf{tune}/\textit{tube} was quite masterfully \textbf{played by the pianist}/\textit{installed by the plumber}. \\
The doctor sees patients with \textbf{mental}/\textit{metal} or other kinds of \textbf{illnesses affecting them}/\textit{fragments in their bodies}. \\
With two strong \textbf{wings}/\textit{wins} this very powerful \textbf{bird can fly for hours}/\textit{team can qualify for the finals}. \\
These \textbf{bills}/\textit{hills} definitely appear to be \textbf{counterfeit, unfortunately}/\textit{steep, unfortunately}. \\
The \textbf{facts}/\textit{faces} and other details of the \textbf{case were published}/\textit{people were blurred}. \\
The rules state that the \textbf{loser}/\textit{lower} of the two relevant \textbf{matches would be eliminated}/\textit{amounts would be considered}. \\
The \textbf{garbage}/\textit{garage} definitely needs to be \textbf{tossed out due to its smell}/\textit{renovated due to its leaks}. \\
This \textbf{wind}/\textit{wine} is most definitely \textbf{blowing from the west}/\textit{delicious with steak}. \\
Thank you for \textbf{taking}/\textit{making} such good \textbf{care of her yesterday}/\textit{pancakes for breakfast}. \\
After being \textbf{seated}/\textit{sealed} carefully in the small \textbf{theater, the audience fell silent}/\textit{jar, the jam was sold}. \\
Because of all the \textbf{treats}/\textit{threats} my dog has become very \textbf{overweight and unhealthy}/\textit{scared and timid}. \\
The \textbf{chart}/\textit{cart} contained lots of \textbf{data and statistics}/\textit{fruits and vegetables}. \\
These \textbf{saints}/\textit{stains} had been talked about and \textbf{venerated by the worshippers}/\textit{removed by the cleaners}. \\
She knew how to \textbf{untie}/\textit{unite} every single one of the \textbf{knots for the sailboat}/\textit{community members}. \\
Are you able to \textbf{reverse}/\textit{reserve} both of these \textbf{transactions over the phone}/\textit{rooms in the hotel}? \\
The documents were \textbf{stored}/\textit{sorted} by the intern in \textbf{filing cabinets}/\textit{alphabetical order}. \\
He was \textbf{tired}/\textit{tried} a long while after his \textbf{shift ended}/\textit{arrest and indictment}. \\
My friend is \textbf{carving}/\textit{craving} a massive \textbf{marble statue}/\textit{cheeseburger and fries}. \\
She \textbf{fried}/\textit{fired} every single one of the \textbf{chicken nuggets}/\textit{employees and consultants}. \\
The \textbf{trial}/\textit{trail} that was located in the \textbf{courthouse attracted media attention}/\textit{woods attracted hikers and campers}. \\
Unlike the \textbf{males}/\textit{meals} we discovered that the \textbf{females have dull feathers}/\textit{transportation was not reimbursed}. \\
The chef's boss said his \textbf{bread}/\textit{beard} definitely needs to be \textbf{baked at four hundred degrees}/\textit{shaved off completely}. \\
These \textbf{genes}/\textit{jeans} can be quickly and cheaply \textbf{sequenced by a lab}/\textit{washed at the laundromat}. \\
\hline
\end{supertabular}

\section{Computational Modeling Details}
\label{sec:model-details}

The noisy-channel model of \citet{clarkModelApproximateIncremental2025, clarkResourceRationalNoisyChannelLanguage2025} was used to generate surprisal values and posterior word error probabilities for all items in the study. 
The following parameters were used: \verb|num_particles| = 128, \verb|conditional_rejuv| = False, \verb|second_pass_rejuv| = True, \verb|second_pass_rejuv_p| = 1.0, \verb|second_pass_rejuv_iters| = 3, \verb|lm_method| = \verb|gpt2|, \verb|normal_alpha| = 3, \verb|error_alpha| = 1, \verb|prompt| = 1.  
A restricted vocabulary formed from the union of the top 5000 most frequent English words, according to the SUBTLEX-US dataset \citep{brysbaertMovingKuceraFrancis2009} and all words appearing in the experimental materials was used. 
All other parameters were set at the default values.

\begin{figure*}[htb]
    \centering
    \includegraphics[width=0.9\linewidth]{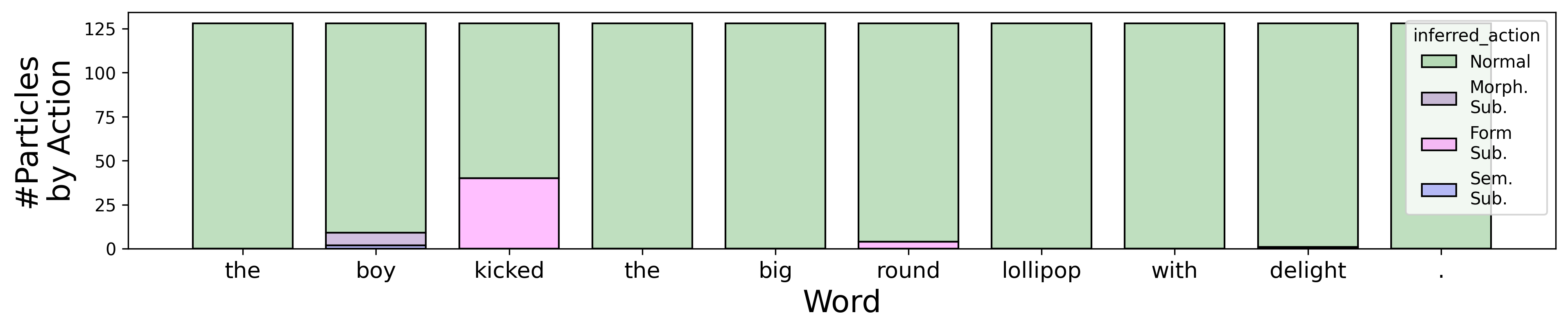}
    \caption{Posterior over actions for example sentence.}
    \label{fig:actions-posterior-example}
\end{figure*}

\begin{figure*}[htb]
    \centering
    \includegraphics[width=0.75\linewidth]{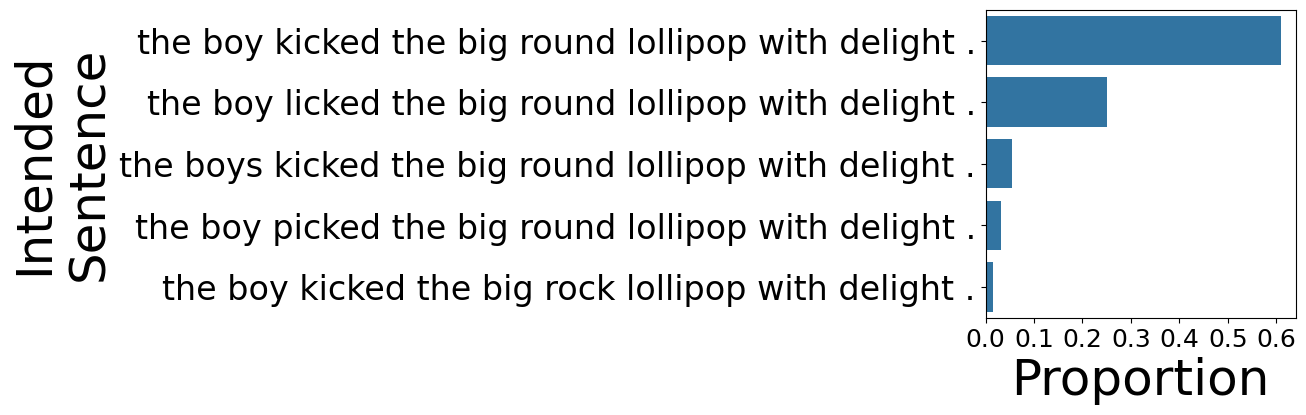}
    \caption{Posterior over inferred intended messages for example sentence.}
    \label{fig:latent-sentence-posterior-example}
\end{figure*}

\begin{figure*}[htb]
    \centering
    \includegraphics[width=0.75\linewidth]{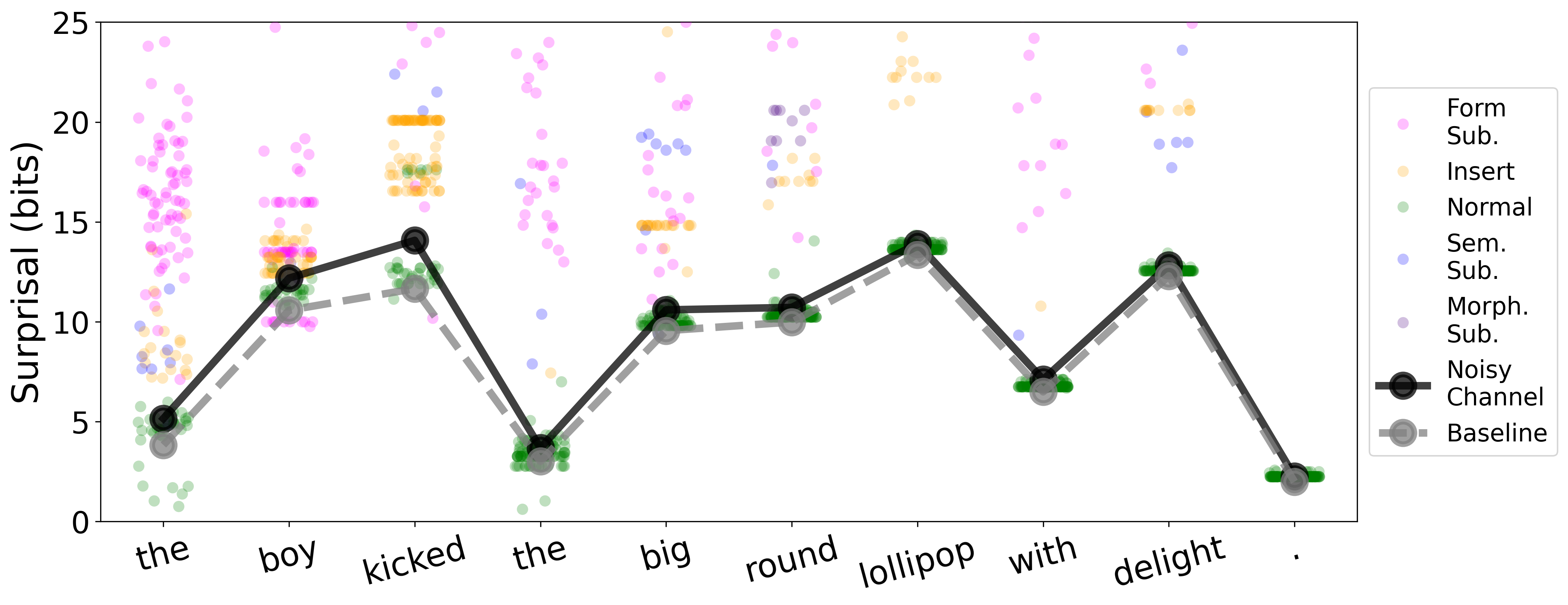}
    \caption{Per-word surprisal under baseline and noisy-channel language models.}
    \label{fig:surprisals-example}
\end{figure*}

\begin{figure*}[htb]
    \centering
    \includegraphics[width=0.9\linewidth]{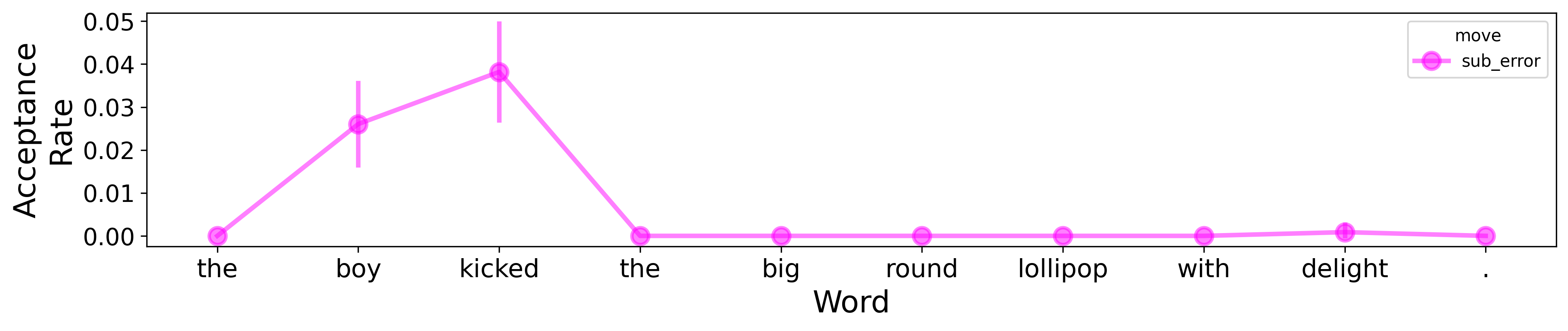}
    \caption{Rejuvenation acceptance rate at each word position for example sentence.}
    \label{fig:rejuvenation-example}
\end{figure*}

\Cref{fig:actions-posterior-example} shows the posterior distribution over actions at each word for an example sentence, extracted from the final set of 128 particles at the end of an inference run. 
\Cref{fig:latent-sentence-posterior-example} shows the posterior distribution over inferred intended sentences for the same example sentence. The literal sentence is still the most likely candidate, but the alternative sentence with \textit{licked} substituted for \textit{kicked} is the next most likely candidate. 
\Cref{fig:surprisals-example} shows the noisy-channel model's surprisal at each observation, with a comparison to baseline language model surprisal from GPT-2 (using the locally-constrained decoding method described in \citet{clarkResourceRationalNoisyChannelLanguage2025} to enforce a restricted vocabulary). 
\Cref{fig:rejuvenation-example} shows the per-word acceptance rate for second-pass rejuvenations. Rejuvenations are most likely at words which have near neighbors that would make the sentence \textit{a priori} more plausible.

We note that the ``noisy-channel garden path'' sentences provide a distinct challenge for computational models of language processing, and provide a direct contrast with incremental noisy-channel processing in sentences like \textit{The storyteller could turn any story into an amusing antidote} \citep{ryskinERPIndexRealtime2021}, as modeled in \citep{clarkResourceRationalNoisyChannelLanguage2025}. In those sentences, the anomalous word always occurred sentence-finally, and any error correction done by inferential comprehenders could be performed immediately upon observing the anomalous word. A key finding was that for such sentences, the surprisal on the final anomalous word would be lower under the noisy-channel model than under a baseline language model, due to the ability to recover an intended alternative (e.g. \textit{anecdote}). 
\Cref{fig:surprisals-example} show that no such reduction in incremental surprisal occurs at the anomalous predicate for this example, illustrating a ``garden-path''-like effect. In contrast, rejuvenation moves for earlier choices enable the processor to arrive at inferences about intended alternatives. 

\FloatBarrier

\section{Statistical Analysis: Full Results}
\label{sec:stats-results}

\begin{minipage}{\columnwidth}
    \captionof{table}{\label{tab:tab:coefficients_RegOutLogistic_simple}Fixed Effects Coefficients with \\ 95\% Credible Intervals (Regressions Out)}
    \centering
    \scriptsize
\begin{tabular}[htb]{lrc}
\toprule
Coefficient & Estimate & 95\% CI\\
\midrule
\textbf{Intercept} & \textbf{-3.883} & \textbf{[-4.219, -3.550]}\\
\textbf{log\_freq} & \textbf{0.06} & \textbf{[0.035, 0.085]}\\
\textbf{word\_nchar} & \textbf{0.08} & \textbf{[0.051, 0.109]}\\
\textbf{surprisal} & \textbf{0.055} & \textbf{[0.036, 0.073]}\\
\textbf{word\_num\_in\_sent} & \textbf{-0.074} & \textbf{[-0.103, -0.046]}\\
pos\_tagFUNCTION & -0.1 & {}[-0.213, 0.012]\\
pos\_tagADJ & -0.001 & {}[-0.110, 0.110]\\
pos\_tagVERB & 0.048 & {}[-0.047, 0.143]\\
pos\_tagADV & -0.021 & {}[-0.144, 0.101]\\
Preamble & -0.032 & {}[-0.164, 0.102]\\
CriticalWord & 0.073 & {}[-0.058, 0.205]\\
\textbf{Predicate} & \textbf{0.309} & \textbf{[0.183, 0.435]}\\
\NeighborGP{} & 0.026 & {}[-0.081, 0.133]\\
\textbf{Typo} & \textbf{0.125} & \textbf{[0.015, 0.231]}\\
\UnrelatedGP{} & -0.011 & {}[-0.116, 0.096]\\
\LateError{} & 0.055 & {}[-0.049, 0.160]\\
Preamble $\times$ \NeighborGP{} & 0.045 & {}[-0.114, 0.196]\\
CriticalWord $\times$ \NeighborGP{} & -0.056 & {}[-0.212, 0.106]\\
\textbf{Predicate $\times$ \NeighborGP} & \textbf{0.333} & \textbf{[0.196, 0.469]}\\
Preamble $\times$ \Typo{} & -0.054 & {}[-0.214, 0.102]\\
CriticalWord $\times$ \Typo{} & 0.001 & {}[-0.202, 0.198]\\
\textbf{Predicate $\times$ \Typo{}} & \textbf{-0.45} & \textbf{[-0.597, -0.300]}\\
Preamble $\times$ \UnrelatedGP{} & -0.013 & {}[-0.176, 0.144]\\
CriticalWord $\times$ \UnrelatedGP{} & 0.059 & {}[-0.094, 0.218]\\
\textbf{Predicate $\times$ \UnrelatedGP{}} & \textbf{0.23} & \textbf{[0.091, 0.368]}\\
Preamble $\times$ \LateError{} & -0.012 & {}[-0.171, 0.145]\\
CriticalWord $\times$ \LateError{} & 0.024 & {}[-0.135, 0.180]\\
\textbf{Predicate $\times$ \LateError{}} & \textbf{0.166} & \textbf{[0.028, 0.305]}\\
\bottomrule
\end{tabular}
\end{minipage}

\vspace{0.5em}  

\begin{minipage}{\columnwidth}
    \captionof{table}{\label{tab:tab:coefficients_RegInLogistic_simple}Fixed Effects Coefficients with \\ 95\% Credible Intervals (Regessions In)}
    \centering
    \scriptsize
\begin{tabular}[htb]{lrc}
\toprule
Coefficient & Estimate & 95\% CI\\
\midrule
\textbf{Intercept} & \textbf{-2.359} & \textbf{[-2.631, -2.083]}\\
\textbf{log\_freq} & \textbf{-0.021} & \textbf{[-0.040, -0.003]}\\
\textbf{word\_nchar} & \textbf{0.178} & \textbf{[0.159, 0.197]}\\
\textbf{surprisal} & \textbf{0.036} & \textbf{[0.025, 0.047]}\\
\textbf{word\_num\_in\_sent} & \textbf{-0.134} & \textbf{[-0.158, -0.110]}\\
\textbf{pos\_tagFUNCTION} & \textbf{-0.225} & \textbf{[-0.311, -0.140]}\\
\textbf{pos\_tagADJ} & \textbf{0.103} & \textbf{[0.021, 0.183]}\\
\textbf{pos\_tagVERB} & \textbf{0.096} & \textbf{[0.031, 0.163]}\\
\textbf{pos\_tagADV} & \textbf{-0.138} & \textbf{[-0.235, -0.042]}\\
\textbf{Preamble} & \textbf{-0.199} & \textbf{[-0.313, -0.080]}\\
\textbf{CriticalWord} & \textbf{0.177} & \textbf{[0.069, 0.279]}\\
\textbf{Predicate} & \textbf{-0.187} & \textbf{[-0.294, -0.082]}\\
\textbf{\NeighborGP} & \textbf{0.296} & \textbf{[0.218, 0.376]}\\
\textbf{\Typo{}} & \textbf{-1.244} & \textbf{[-1.354, -1.136]}\\
\textbf{\UnrelatedGP{}} & \textbf{0.389} & \textbf{[0.309, 0.468]}\\
\textbf{\LateError{}} & \textbf{0.452} & \textbf{[0.376, 0.530]}\\
Preamble $\times$ \NeighborGP{} & 0.035 & {}[-0.097, 0.165]\\
\textbf{CriticalWord $\times$ \NeighborGP} & \textbf{0.369} & \textbf{[0.242, 0.499]}\\
\textbf{Predicate $\times$ \NeighborGP} & \textbf{-0.134} & \textbf{[-0.258, -0.012]}\\
Preamble $\times$ \Typo{} & -0.047 & {}[-0.208, 0.116]\\
CriticalWord $\times$ \Typo{} & 0 & {}[-0.202, 0.189]\\
\textbf{Predicate $\times$ \Typo{}} & \textbf{-0.315} & \textbf{[-0.466, -0.161]}\\
Preamble $\times$ \UnrelatedGP{} & 0.085 & {}[-0.043, 0.216]\\
\textbf{CriticalWord $\times$ \UnrelatedGP{}} & \textbf{0.279} & \textbf{[0.156, 0.406]}\\
\textbf{Predicate $\times$ \UnrelatedGP{}} & \textbf{-0.138} & \textbf{[-0.261, -0.018]}\\
Preamble $\times$ \LateError{} & -0.117 & {}[-0.242, 0.011]\\
CriticalWord $\times$ \LateError{} & 0.005 & {}[-0.118, 0.127]\\
Predicate $\times$ \LateError{} & -0.043 & {}[-0.168, 0.080]\\
\bottomrule
\end{tabular}
\end{minipage}

\vspace{0.5em}

\begin{minipage}{\columnwidth}
    \captionof{table}{\label{tab:tab:coefficients_simple}Fixed Effects Coefficients with \\ 95\% Credible Intervals (Re-reading Time)}
    \centering
    \scriptsize
    \begin{tabular}[htb]{lrc}
\toprule
Coefficient & Estimate & 95\% CI\\
\midrule
\midrule
\textbf{Intercept} & \textbf{6.077} & \textbf{[6.002, 6.146]}\\
\textbf{hu\_Intercept} & \textbf{0.726} & \textbf{[0.676, 0.775]}\\
log\_freq & -0.011 & {}[-0.039, 0.016]\\
\textbf{word\_nchar} & \textbf{0.156} & \textbf{[0.138, 0.173]}\\
surprisal\_nc & 0.016 & {}[-0.000, 0.033]\\
word\_num\_in\_sent & -0.006 & {}[-0.016, 0.004]\\
\textbf{pos\_tagFUNCTION} & \textbf{0.043} & \textbf{[0.006, 0.081]}\\
pos\_tagADJ & 0.014 & {}[-0.024, 0.051]\\
pos\_tagVERB & 0.021 & {}[-0.012, 0.053]\\
pos\_tagADV & -0.039 & {}[-0.088, 0.010]\\
\textbf{Preamble} & \textbf{0.106} & \textbf{[0.038, 0.174]}\\
CriticalWord & 0.074 & {}[-0.007, 0.152]\\
\textbf{Predicate} & \textbf{0.093} & \textbf{[0.022, 0.160]}\\
\textbf{\NeighborGP} & \textbf{-0.055} & \textbf{[-0.105, -0.003]}\\
\textbf{\Typo{}} & \textbf{-0.148} & \textbf{[-0.226, -0.071]}\\
\UnrelatedGP{} & -0.028 & {}[-0.080, 0.022]\\
\textbf{\LateError{}} & \textbf{-0.055} & \textbf{[-0.106, -0.004]}\\
Preamble $\times$ \NeighborGP{} & -0.022 & {}[-0.100, 0.053]\\
\textbf{CriticalWord $\times$ \NeighborGP} & \textbf{0.215} & \textbf{[0.125, 0.310]}\\
Predicate $\times$ \NeighborGP{} & -0.039 & {}[-0.120, 0.043]\\
Preamble $\times$ \Typo{} & 0.088 & {}[-0.018, 0.193]\\
\textbf{CriticalWord $\times$ \Typo{}} & \textbf{0.456} & \textbf{[0.333, 0.578]}\\
Predicate $\times$ \Typo{} & 0.043 & {}[-0.099, 0.184]\\
Preamble $\times$ \UnrelatedGP{} & -0.065 & {}[-0.142, 0.011]\\
\textbf{CriticalWord $\times$ \UnrelatedGP{}} & \textbf{0.101} & \textbf{[0.007, 0.196]}\\
Predicate $\times$ \UnrelatedGP{} & -0.071 & {}[-0.150, 0.013]\\
Preamble $\times$ \LateError{} & -0.015 & {}[-0.096, 0.060]\\
\textbf{CriticalWord $\times$ \LateError{}} & \textbf{0.133} & \textbf{[0.037, 0.232]}\\
Predicate $\times$ \LateError{} & 0.021 & {}[-0.063, 0.104]\\
\textbf{hu\_log\_freq} & \textbf{0.078} & \textbf{[0.042, 0.114]}\\
\textbf{hu\_word\_nchar} & \textbf{-0.19} & \textbf{[-0.216, -0.164]}\\
\textbf{hu\_surprisal\_nc} & \textbf{-0.167} & \textbf{[-0.195, -0.141]}\\
\textbf{hu\_word\_num\_in\_sent} & \textbf{0.138} & \textbf{[0.132, 0.144]}\\
\textbf{hu\_pos\_tagFUNCTION} & \textbf{-0.067} & \textbf{[-0.130, -0.005]}\\
\textbf{hu\_pos\_tagADJ} & \textbf{-0.147} & \textbf{[-0.210, -0.085]}\\
\textbf{hu\_pos\_tagVERB} & \textbf{-0.168} & \textbf{[-0.221, -0.111]}\\
hu\_pos\_tagADV & 0.025 & {}[-0.055, 0.104]\\
\bottomrule
\end{tabular}
\end{minipage}

\clearpage

\section{Pointwise Mutual Information}
\label{sec:appendix-pmi}

In a post-hoc anlaysis, we quantify the associative strength between a \CriticalWord{} and its
corresponding \Predicate{} using pointwise mutual information (PMI) estimated
from a masked language model, following past work \citep{wilcoxInformationtheoreticAnalysisTargeted2024, hooverLinguisticDependenciesStatistical2021a}.
All estimates were obtained from RoBERTa-large
\citep{liuRoBERTaRobustlyOptimized2019a}, a bidirectional transformer pretrained on approximately 160\,GB of English text.

\subsection*{PMI estimation}

For a given variant ``$\dots$ [\CriticalWord{}=$c$] $\dots$ [\Predicate{}=$p$]'' we estimate $\mathrm{PMI}(c\,;\,p)$ as
\[
  \log_2 \frac{P(c \mid p,\,\text{context})}{P(c \mid \text{context})}
\]
using two masked-prediction queries to RoBERTa-large.

\paragraph{Numerator.}
The critical word position is replaced by \texttt{[MASK]} while the
predicate phrase is left intact: For example, the sentence \textit{The boy licked the big round lollipop with delight} would yield \textit{The boy [MASK] the big round lollipop with delight}. 
The model's predicted probability at that mask position gives
$P(c \mid p, \text{context})$.

\paragraph{Denominator.}
Both the critical word and the entire predicate phrase are
replaced by mask tokens.  The predicate phrase may span $k \geq 1$
subword tokens; it is replaced by exactly $k$ consecutive \texttt{[MASK]}
tokens (written $\texttt{[MASK]}^k$) to preserve sentence length:
$S_{\mathrm{den}} \;=\; \Preamble{} \;\; \texttt{[MASK]} \;\; \Intervening{} \;\; \texttt{[MASK]}^k.$
The model's predicted probability at the critical-word mask gives $P(c \mid \text{context})$.

\paragraph{PMI score.}
\begin{align*}
  \mathrm{PMI}(c;\,p)
    &\;=\; \log_2 P(c \mid S_{\mathrm{num}}) \\
    &\quad\; -\; \log_2 P(c \mid S_{\mathrm{den}}),
\end{align*}
expressed in bits.  A positive value indicates that the predicate raises
the model's probability of the critical word above its baseline. 
Each item yields two \Plausible{} variants and two \NeighborGP{} variants, using a total of two unique \CriticalWord{} values. 
Items for which either value of \CriticalWord{} tokenizes to more than one subword token are excluded from this analysis. Only three items were excluded on this basis. 

\subsection*{Per-item aggregate}

Let $\pi_{jk}^{(i)} \equiv \mathrm{PMI}(c_j^{(i)};\,p_k^{(i)})$, where $j,k \in \{1,2\}$ index the two variants of each item in the \Plausible{} and \NeighborGP{} conditions (e.g., \textit{licked}/\textit{kicked} and \textit{ball into the net}/\textit{lollipop with delight}).
Item-level condition means are formed by averaging the two
within-condition scores:
\begin{align*}
  \widehat{\mathrm{PMI}}_{\mathrm{\Plausible{}}}^{(i)}
    &\;=\; \tfrac{1}{2}\!\left[p_{11}^{(i)} + p_{22}^{(i)}\right], \\[4pt]
  \widehat{\mathrm{PMI}}_{\mathrm{\NeighborGP{}}}^{(i)}
    &\;=\; \tfrac{1}{2}\!\left[p_{12}^{(i)} + p_{21}^{(i)}\right].
\end{align*}
Statistical comparison uses a Wilcoxon signed-rank test on the
per-item differences
\[
  \Delta^{(i)}
    \;=\; \widehat{\mathrm{PMI}}_{\mathrm{\Plausible{}}}^{(i)}
        - \widehat{\mathrm{PMI}}_{\mathrm{\NeighborGP{}}}^{(i)},
\]
demonstrating that the two distributions of PMI values are almost certainly drawn from different distributions ($p < 0.001$). 

\begin{figure}[htb]
    \centering
    \includegraphics[width=\linewidth]{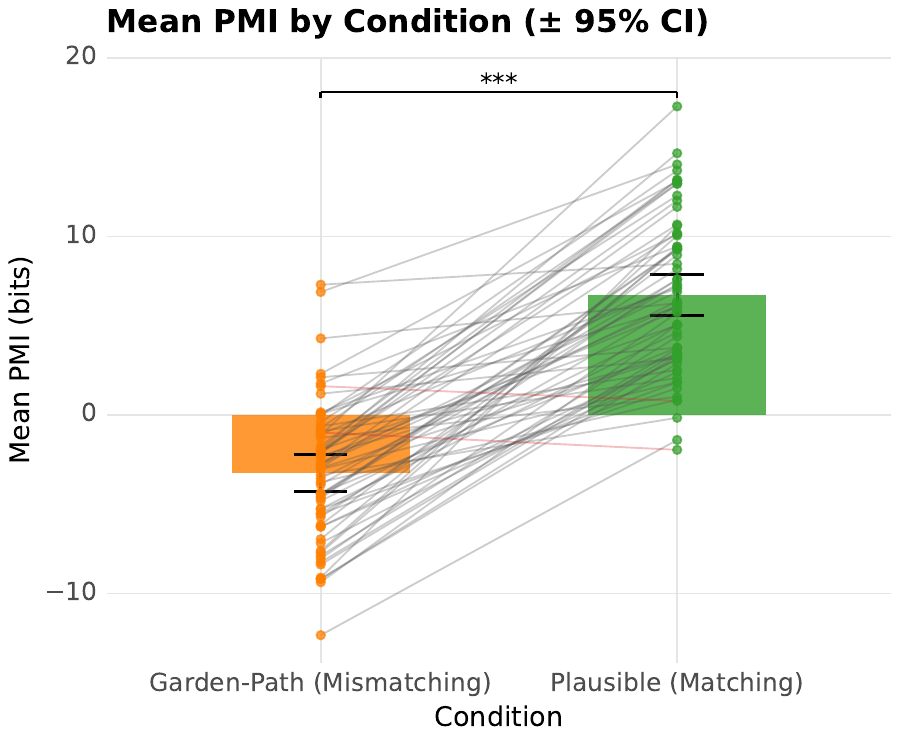}
    \caption{Pointwise mutual information across items and conditions (\Plausible{} vs. \NeighborGP{}) in the experimental materials. Lines connect values of $\pi_{11}^{(i)}$ and $\pi_{12}^{(i)}$, and connect values of $\pi_{22}^{(i)}$ and $\pi_{21}^{(i)}$. For all except two pairs, the \Plausible{} condition has higher PMI than the \NeighborGP{} condition.}
    \label{fig:pmi-barplot}
\end{figure}

\end{document}